This manuscript has been accepted for publication in IEEE Transactions on Systems, Man, and Cybernetics: Systems.

# Differential Flatness-based Fast Trajectory Planning for Fixed-wing Unmanned Aerial Vehicles

Junzhi Li, Jingliang Sun, Teng Long and Zhenlin Zhou

***Abstract*—Due to the strong nonlinearity and nonholonomic dynamics, despite the various general trajectory optimization methods presented, few of them can guarantee efficient computation and physical feasibility for relatively complicated fixed-wing UAV dynamics. Aiming at this issue, this paper investigates a differential flatness-based trajectory optimization method for fixed-wing UAVs (DFTO-FW). The customized trajectory representation is presented through differential flat characteristics analysis and polynomial parameterization, eliminating equality constraints to avoid the heavy computational burdens of solving complex dynamics. Through the design of integral performance costs and derivation of analytical gradients, the original trajectory optimization is transcribed into a lightweight, unconstrained, gradient-analytical optimization with linear time complexity to improve efficiency further. The simulation experiments illustrate the superior efficiency of the DFTO-FW, which takes sub-second CPU time (on a personal desktop) against other competitors by orders of magnitude to generate fixed-wing UAV trajectories in randomly generated obstacle environments.**

## I. Introduction

In recent years, unmanned aerial vehicles (UAVs) have been extensively studied in many areas, e.g., search and rescue, coverage reconnaissance, and environmental observation [1]–[4]. Owing to favorable long endurance and high-altitude flight capabilities, fixed-wing UAVs are more attractive to effectively and efficiently perform the aforementioned missions [5]. Trajectory planning generates smooth, dynamically precise, and collision-free motions for fixed-wing UAVs. However, compared with unmanned helicopters and multi-rotor drones, trajectory planning for fixed-wing vehicles suffers from the strong nonlinearity of nonholonomic dynamics (e.g., minimum flight speed, bounded turning curvatures, and coupled control channels). Despite the fact that various general trajectory planning tools have been presented, few of them can guarantee fast convergence and physical feasibility for relatively complicated fixed-wing dynamics. Therefore, achieving efficient fixed-wing UAV trajectory generation is still challenging.

Existing typical trajectory planning approaches with high computation efficiency include graph-based search [6]–[8], sampling-based search [9]–[11], and geometric tangent-based methods [12]–[14], etc. However, these methods mainly use oversimplified kinematics for better computation efficiency while ignoring realistic control constraints (e.g., variable flight speed and limited engine forces). Thus, it is hard for these methods to obtain sufficiently smooth, continuous, traceable trajectories, limiting the full exploitation of UAVs' capability.

Optimization-based methods [15]–[21] formulate trajectory planning as an optimal control problem (OCP) and solve the OCP by numerical optimization. Many general-purpose tools (e.g., the collocation-based toolkit GPOPS-II [15] and the shooting-based ACADO [16]) can solve such OCP to generate smooth trajectories with high dynamics fidelity. However, such OCP is generally a nonlinear, nonconvex, NP-hard problem [17]. These general-purpose methods usually face the heavy computational burdens of directly solving nonlinear optimization problems with complicated dynamics and cannot guarantee fast convergence. For example, Barry *et al*. [18] reported that it takes 3-5 min of computation time to generate a 4.5 m long trajectory for 6-DOF fixed-wing aircraft by using sparse nonlinear optimization (SNOPT). To some extent, sequential convex programming (SCP) methods [19], [20] can alleviate the heavy time consumption by convexification of nonlinear dynamics and sequential decomposition of complex problems. Nevertheless, SCP methods are prone to the infeasibility of dynamics convexification, resulting in frequent inefficient sequential iteration and decreasing efficiency and robustness [22]. Summarily, the above-mentioned methods optimize trajectory with complex dynamics, regardless of the intrinsic dynamic characteristics, and thus struggle with efficiency and solution quality. Hence, developing customized fast trajectory optimization methods for fixed-wing UAVs is an urgent calling.

To achieve efficient trajectory optimization, the concept of differential flatness [23] has become attractive. Through flat dynamics transcription and trajectory representations (e.g., polynomial [24]–[26], Bézier curves [27], [28], and other splines [29]–[31]), one can directly optimize the trajectories that naturally satisfy the system differential dynamics, instead of simply regarding the dynamics as a constraint, which dramatically facilitates efficient trajectory generation. For

**NOTICE**: This manuscript has been accepted for publication in IEEE Transactions on Systems, Man, and Cybernetics: Systems. This is the author's version of the accepted manuscript. Revisions have been made to improve grammar and formatting. The final version will appear in the IEEE Xplore digital library.

platforms like quadrotors and helicopters, due to their relatively simple dynamics and straightforward differential flatness properties, the trajectory planning approaches have achieved significant progress with efficiency and practical applicability. Kumar *et al.* [24], [25] investigated the differential flatness features of quadrotors and used fixed-time polynomials to formulate quadratic programming (QP) problems for minimizing the trajectory snap. However, their work did not address obstacle avoidance. Tordesillas *et al.* [29], [30] proposed outer polyhedral representations for quadrotors and achieved highly efficient collision avoidance in unknown and dynamic environments. Wang *et al.* customized a class of spatial-temporal deformable polynomial splines called MINCO (minimum control) [31]. They proposed a gradient-analytical unconstrained trajectory optimization framework to achieve millisecond-level online trajectory generation for multicopters [32]–[34]. Duan *et al.* [35] investigated the trajectory optimization for multi-helicopter cooperative transportation by MINCO expression. However, compared with these simple platforms, trajectory optimization for fixed-wing UAVs presents unique challenges regarding strong nonholonomic and coupled controls. Since the nonlinear dynamics are not explicitly differentially flat, to the best of our knowledge, few researchers use flatness-based approaches for trajectory optimization. As a result, typical methods can hardly achieve the same efficiency and robustness as demonstrated in quadrotor applications. Although related flatness-based works such as [26] explored this topic, their approaches are limited to simplified 2D coordinated turn motions with constant speed. [36] focused on the differential flatness of VTOL drones but did not incorporate obstacle avoidance in trajectory optimization. Therefore, customized differential-flatness-based trajectory optimization methods for fixed-wing UAVs are significant for rapid trajectory generation.

Inspired by the preceding discussion, this paper investigates the differential flatness-based trajectory optimization framework for fixed-wing UAVs (DFTO-FW) as shown in Fig.1. Trajectory optimization is formulated as a constrained optimal control problem subject to nonlinear dynamics, obstacle avoidance, and terminal and performance constraints. Then, the differential flatness characteristic of the 3D dynamics of fixed-wing UAVs with variable speed is revealed. The flatness-based trajectory representation for fixed-wing dynamics is presented through uniform-time polynomial parameterization, which eliminates the nonlinear dynamics and avoids the computational burden of satisfying complex dynamic equality constraints in optimization iteration. After that, this paper further transcribes the original problem into a lightweight, unconstrained, gradient-analytical optimization with linear time complexity in each iteration to achieve fast trajectory generation.

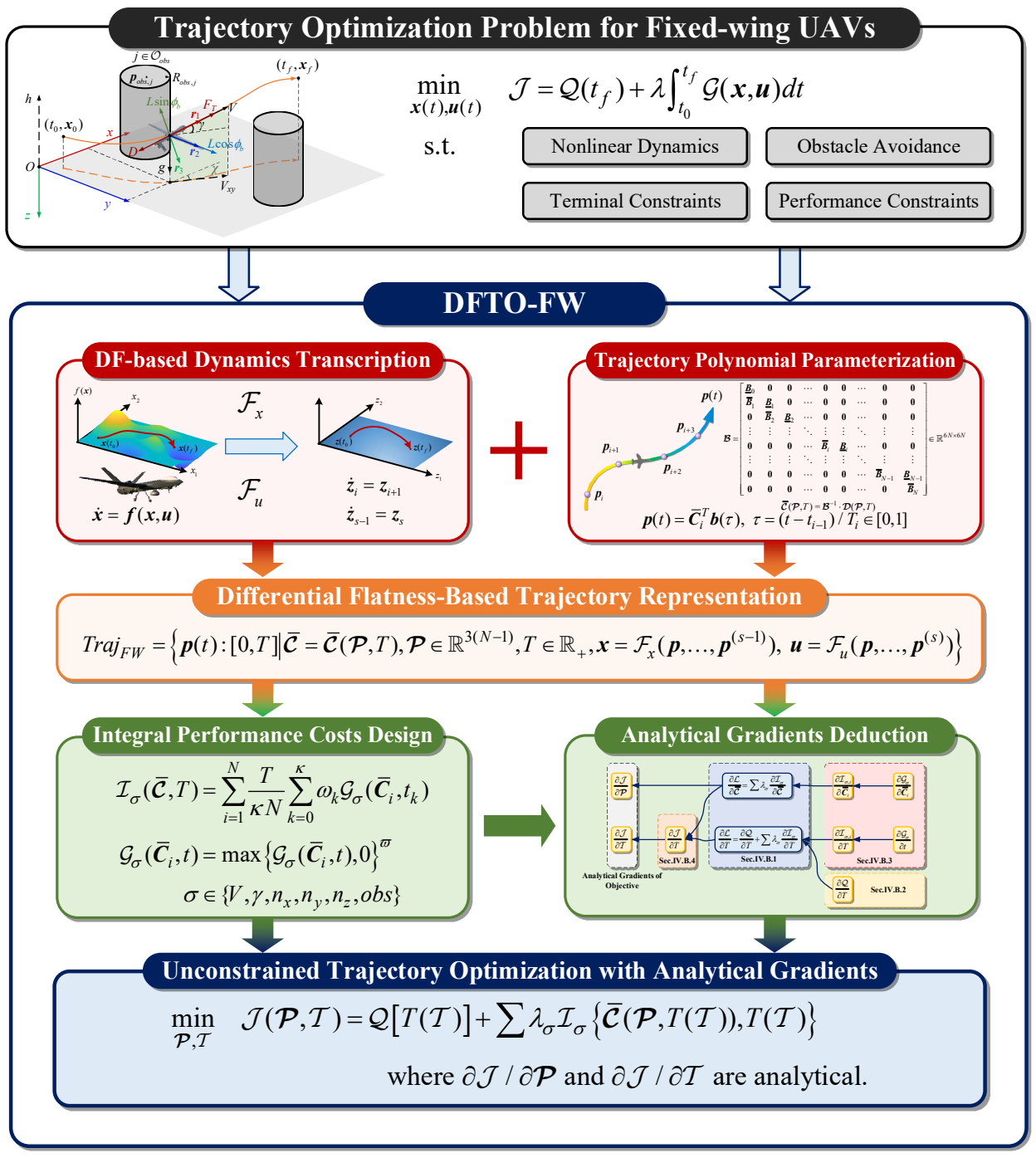

**Fig.1** Overview of the technical framework.

Compared to existing approaches, the key innovations and contributions of this study are as follows.

1) The differential flatness characteristic of the 3D dynamics of fixed-wing UAVs with variable speed is revealed. Unlike existing works [20], which focus solely on simplified 2D coordinated turn motions with constant speed, our model supports variable-speed maneuvers in 3-D space and considers obstacle avoidance, enabling more realistic trajectory planning.
2) A flatness-based trajectory representation model for fixed-wing UAVs is presented. Unlike [19], [20], which use simple linear convexification and are prone to infeasibility and inefficient computation, our approach eliminates the equality constraints. It generates the solution that naturally satisfies UAV dynamics, which avoids the computational burdens and infeasibility of optimization on complex nonlinear dynamics. As a result, our method provides superior efficiency and a higher success rate than competitors [15], [20].
3) A differential flatness-based fast trajectory optimization method for fixed-wing UAVs (DFTO-FW) is proposed. Compared with general-purpose tools that directly solve the OCP, our approach transcribes the problem into a lightweight, unconstrained optimization, significantly reducing computational complexity. Additionally, DFTO-FW derives analytical gradients to accelerate optimization iteration, resulting in linear time complexity per iteration, which allows our method to outperform GPOPS-II [15] and TRF-SCP [20] by orders of magnitude in computation efficiency and dramatically decreases the computation time from $10^{-1}$–$10^{1}$ s to $10^{-2}$–$10^{-1}$ s, satisfying the online solving requirements.

The remainder of this paper is organized as follows. Sec.II presents the problem formulation. Sec.III analyzes the differential flatness characteristics of fixed-wing UAVs and introduces the polynomial-based trajectory representation model. Sec.IV formulates the original problem into an unconstrained optimization problem with analytical gradients to reduce computational costs. Simulation experiments are conducted in Sec.0, followed by the conclusions in Sec.VI.

## II. Problem Formulation and Preliminaries

### A. Problem Formulation

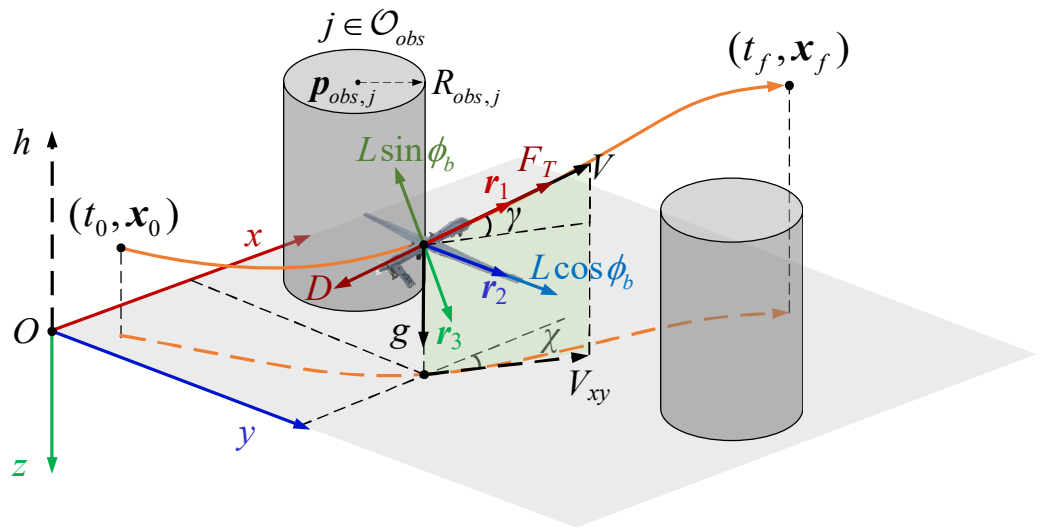

**Fig.2** Scenario of trajectory planning for a fixed-wing UAV.

Fig.2 shows a typical scenario of trajectory planning for a fixed-wing UAV. $Oxyz$ is the inertial frame (north-east-down, NED frame); $x$, $y$ and $z$ denote the coordinates; $h=-z$ is the flight altitude; $\boldsymbol{r}_1$, $\boldsymbol{r}_2$ and $\boldsymbol{r}_3$ are three axes of the flight path frame; $V$ is the flight speed; $\chi$ is the heading angle; $\gamma$ is the flight path angle; $\phi_b$ is the bank angle; $g$ represents the gravitational acceleration; $m$ is the mass; $F_T$ is the engine thrust, $D$ and $L$ are the aerodynamic drag and lift, respectively. Assuming the UAV has a constant mass and time-invariant center of mass, the flight dynamics can be described as follows [37].

$$\dot{\boldsymbol{x}} = \boldsymbol{f}(\boldsymbol{x},\boldsymbol{u}) = \begin{bmatrix} V\cos\gamma\cos\chi \\ V\cos\gamma\sin\chi \\ -V\sin\gamma \\ (F_T - D)/m - g\sin\gamma \\ L\sin\phi_b/(mV\cos\gamma) \\ L\cos\phi_b/(mV) - g\cos\gamma/V \end{bmatrix} \tag{1}$$

in which, $\boldsymbol{x}=[x,y,z,V,\chi,\gamma]^\top \in \mathbb{R}^6$ and $\boldsymbol{u}=[F_T, L, \phi_b]^\top \in \mathbb{R}^3$ are the states and controls, respectively.

Record $t_0$ and $t_f$ as the initial and final times, respectively. The constraints of trajectory planning are given as follows.

1) **Flight performance constraints**: During the flight, the speed is restricted by the constraint $0 < V_{\min} < V \le V_{\max}$. $\gamma$ is limited due to climb/dive rate performance, i.e., $\gamma_{\min} \le \gamma \le \gamma_{\max}$. Besides, the controls are bounded by limited engine thrust and maneuverability. To sum up, the performance constraints are described as (2), where the subscript min and max denote the lower and upper boundaries, respectively.

$$\boldsymbol{x}_{\min} \le \boldsymbol{x}(t) \le \boldsymbol{x}_{\max},\ \boldsymbol{u}_{\min} \le \boldsymbol{u}(t) \le \boldsymbol{u}_{\max} \tag{2}$$

2) **Obstacle avoidance constraints**: The fixed-wing UAV typically detours in the horizontal plane owing to stronger horizontal maneuverability to maintain safe distances from obstacles. Considering cylinder obstacles with infinite height, the obstacle avoidance constraints are described as

$$\left\| \boldsymbol{H}\cdot\boldsymbol{x}(t) - \boldsymbol{p}_{obs,j} \right\| \ge R_{obs,j} + R_{safe},\ \forall j \in \mathcal{O}_{obs} \tag{3}$$

where $\mathcal{O}_{obs}$ is the obstacle set; $\boldsymbol{p}_{obs,j} \in \mathbb{R}^2$ and $R_{obs,j}$ denote the center and the radius of the obstacle $j$; $R_{safe}$ is the safe distance; and $\boldsymbol{H} = [\boldsymbol{I}_{2\times2}, \boldsymbol{0}_{2\times4}] \in \mathbb{R}^{2\times6}$ is the matrix to select $x$, $y$ coordinates.

3) **Terminal constraints**. Define $\boldsymbol{x}_0$, $\boldsymbol{x}_f$ as the initial and final states, and $\boldsymbol{u}_0$, $\boldsymbol{u}_f$ as the initial and final controls. The terminal constraints at $t_0$ and $t_f$ are

$$\boldsymbol{x}(t_0)=\boldsymbol{x}_0,\ \boldsymbol{x}(t_f)=\boldsymbol{x}_f,\ \boldsymbol{u}(t_0)=\boldsymbol{u}_0,\ \boldsymbol{u}(t_f)=\boldsymbol{u}_f \tag{4}$$

Then, the trajectory planning for a fixed-wing UAV can be formulated as the following optimal control problem

$$\textbf{Problem 1}:\quad \begin{aligned} &\min_{\boldsymbol{x}(t),\boldsymbol{u}(t)} \ \mathcal{J} = \mathcal{Q}(t_f) + \lambda\int_{t_0}^{t_f} \mathcal{G}(\boldsymbol{x},\boldsymbol{u})dt \\ &\text{s.t.}\quad (1), (2), (3) \text{ and } (4) \end{aligned} \tag{5}$$

where $\mathcal{J}$ is the objective function consists of the time-related $\mathcal{Q}(t_f)$ and integral costs $\int_{t_0}^{t_f}\mathcal{G}(\boldsymbol{x},\boldsymbol{u})dt$; and $\lambda$ is the weight.

**Remark 1**. For practice applications of fixed-wing UAVs, this paper takes $\mathcal{Q}=\|t_f - t_0\|$ (minimizing the flight time) and $\mathcal{G}=\boldsymbol{u}^\top\boldsymbol{u}$ (minimizing control efforts) as examples in the subsequent formula deduction and simulation. However, the design of $\mathcal{J}$ is not limited to the specific form described in this paper. One can design user-defined objective functions for different practice requirements, such as $\mathcal{Q}=\|t_f - t_f^{\text{goal}}\|$ (minimizing arrival time errors), $\mathcal{Q}=\|\boldsymbol{x}_f - \boldsymbol{x}_f^{\text{goal}}\|$ (minimizing goal state errors), and $\mathcal{G}=\|\boldsymbol{x}-\boldsymbol{x}^{\text{ref}}\|^2$ (tracking a reference signal $\boldsymbol{x}^{\text{ref}}$). Indeed, our approach uses analytical gradients to accelerate trajectory optimization. Thus, the objective should be designed in easily differentiated forms, such as norms and quadratics.

**Remark 2**. Compared to simple platforms (e.g., quadrotors or ground robots), the trajectory planning problem shown in (5) presents unique challenges in terms of:

1) **Implicit Differential Flatness**. Unlike the quadrotors with explicitly identifiable flat outputs, the differential flatness of fixed-wing UAVs is not straightforward. This implicit nature complicates the process of parameterizing states and controls, making spline-based representations challenging to apply. Consequently, existing methods often directly solve complex dynamic equality constraints, resulting in inefficient convergence and reduced feasibility.

2) **Strong Nonholonomic**. Fixed-wing vehicles require continuous forward speed to generate lift, and cannot hover, maneuver laterally, and perform sharp turns like quadrotors. Their motions are constrained by bounded forward velocity, turning radius, and climb/dive rate, resulting in more limited feasible regions in trajectory optimization.

3) **Coupled Controls.** The longitudinal, lateral, and vertical motion channels of fixed-wing dynamics are strongly coupled. For instance, changes in engine thrust affect both forward velocity and climb/dive rates. Meanwhile, bank angle influences yaw and lateral motion. The coupled controls increase the complexity of the control and trajectory optimization process.

Due to these challenges, although various general-purpose optimization tools have been presented (e.g., GPOPS-II [15] and ACADO [16]), few of them can guarantee highly efficient solving. Based on differential flat theory, this paper takes a perspective to transcribe the original optimization problem into a lightweight, unconstrained optimization with analytical gradients to achieve fast trajectory generation. The details are shown in subsequent sections.

### *B. Differential Flatness Theory*

**Definition 1** [23]. For a dynamics system $\dot{\boldsymbol{x}} = \boldsymbol{f}(\boldsymbol{x},\boldsymbol{u})$ with states $\boldsymbol{x} \in \mathbb{R}^n$ and controls $\boldsymbol{u} \in \mathbb{R}^m$, it is differential flat when there exist flat outputs $z \in \mathbb{R}^m$, which can parameterize $\boldsymbol{x}$, $\boldsymbol{u}$ with $z$ and finite-order derivatives of $z$ as

$$\begin{aligned} \boldsymbol{x} &= \mathcal{F}_x(z,\dot{z},\ldots,z^{(s-1)}) \\ \boldsymbol{u} &= \mathcal{F}_u(z,\dot{z},\ldots,z^{(s)}) \end{aligned} \tag{6}$$

where $z^{(s)}$ is the $s^{\text{th}}$-order derivative of $z$; $\mathcal{F}_x : \mathbb{R}^{ms} \to \mathbb{R}^n$ and $\mathcal{F}_u : \mathbb{R}^{m(s+1)} \to \mathbb{R}^m$ are flat mappings determined by dynamics $\boldsymbol{f} : \mathbb{R}^{m+n} \to \mathbb{R}^n$. Then, the original system can be accurately linearized into a decoupled integral system as follows.

$$\begin{aligned} \dot{z}^{(i)} &= z^{(i+1)},\ \ i = 0,1,2,\ldots,s-2 \\ \dot{z}^{(s-1)} &= z^{(s)} \end{aligned} \tag{7}$$

For complex nonlinear dynamics, the differential flatness allows trajectory optimization transcript into solving flat outputs $z$ for simple linear integral dynamics. Once $z$ are solved, one can accurately determine the original states $\boldsymbol{x}$ and controls $\boldsymbol{u}$ by flat mappings $\mathcal{F}_x$, $\mathcal{F}_u$. Especially for a system whose flat outputs $z$ are significantly less than the $\boldsymbol{x}$, $\boldsymbol{u}$ in dimensionality, this property can dramatically reduce the optimization dimension and improve efficiency [31].

## III. Differential Flatness-based Trajectory Representation

This section presents the differential flatness-based trajectory representation for fixed-wing UAVs. Firstly, we reveal the differential flatness of fixed-wing dynamics and transcribe the nonlinear dynamics into an integral flat system. Then, via uniform-time polynomial parameterization, the differential flatness-based trajectory representation is presented, which eliminates the dynamic and terminal equality constraints. Finally, the original problem is transformed into solving flat outputs on integral dynamics with lower dimensions, which avoids the heavy computational burdens of directly solving complex dynamics. The details are as follows.

### *A. Differential Flatness-based Dynamics Transcription*

According to **Definition 1**, we select the position vector $\boldsymbol{p} = [x,y,z]^\top \in \mathbb{R}^3$ as the flat output for the fixed-wing dynamics. The 1st-order and 2nd-order derivatives of $\boldsymbol{p}$ are recorded as the velocity $\dot{\boldsymbol{p}} = \boldsymbol{v} = [v_x,v_y,v_z]^\top \in \mathbb{R}^3$ and acceleration $\ddot{\boldsymbol{p}} = \boldsymbol{a} = [a_x,a_y,a_z]^\top \in \mathbb{R}^3$ in the inertial frame, respectively. The flat outputs $\boldsymbol{p}$ represent the translation motion of the fixed-wing vehicle, which provides significant advantages for subsequent spatial-temporal trajectory planning with complex constraints.

As shown in Fig.2, record $\boldsymbol{e}_1 = [1,0,0]^\top$, $\boldsymbol{e}_2 = [0,1,0]^\top$, $\boldsymbol{e}_3 = [0,0,1]^\top$ as the unit vectors of the inertial frame, and $\boldsymbol{r}_1$, $\boldsymbol{r}_2$, $\boldsymbol{r}_3$ as the unit vectors of the flight path frame. $\boldsymbol{r}_1 = \boldsymbol{v} / \|\boldsymbol{v}\|$ parallels the velocity direction. $\boldsymbol{r}_2$ is the vector perpendicular to the velocity in the horizontal plane, and thus $\boldsymbol{r}_2 = (\boldsymbol{e}_3 \times \boldsymbol{v}) / \|\boldsymbol{e}_3 \times \boldsymbol{v}\|$. $\boldsymbol{r}_3$ is determined by the right-hand rule of $\boldsymbol{r}_1$ and $\boldsymbol{r}_2$, thus $\boldsymbol{r}_3 = \boldsymbol{v} \times (\boldsymbol{e}_3 \times \boldsymbol{v}) / \|\boldsymbol{v} \times (\boldsymbol{e}_3 \times \boldsymbol{v})\|$. Then, the flat mappings $\mathcal{F}_x : (\boldsymbol{p},\dot{\boldsymbol{p}}) \to \boldsymbol{x}$ and $\mathcal{F}_u : (\boldsymbol{p},\dot{\boldsymbol{p}},\ddot{\boldsymbol{p}}) \to \boldsymbol{u}$ are derived as follows to transcribe the nonlinear dynamics (1) into a differential flat system.

The flat mapping of states $\mathcal{F}_x : (\boldsymbol{p},\dot{\boldsymbol{p}}) \to \boldsymbol{x}$ describes the relationship from $(\boldsymbol{p},\dot{\boldsymbol{p}})$ to system states $\boldsymbol{x}$.

1) **Cartesian coordinates**. Since $\boldsymbol{p}$ represents the position vector, the coordinates can be computed using the dot product of $\boldsymbol{p}$ with the axes of the inertial frame as

$$x = \boldsymbol{e}_1^\top \boldsymbol{p},\ y = \boldsymbol{e}_2^\top \boldsymbol{p},\ z = \boldsymbol{e}_3^\top \boldsymbol{p} \tag{8}$$

2) **Flight speed**. $V$ is the norm of the velocity vector $\boldsymbol{v}$ as

$$V = \|\boldsymbol{v}\| = (\boldsymbol{v}^\top \boldsymbol{v})^{1/2} \tag{9}$$

3) **Heading angle**. $\chi$ is the angle between $\boldsymbol{v}$ and $\boldsymbol{e}_1$ in the horizontal plane, and can be calculated by

$$\chi = \operatorname{arctan2}(v_y, v_x) = \operatorname{arctan2}(\boldsymbol{e}_2^\top \boldsymbol{v}, \boldsymbol{e}_1^\top \boldsymbol{v}) \tag{10}$$

in which $\operatorname{arctan2}(\cdot)$ is the four-quadrant inverse tangent.

4) **Flight path angle**. $\gamma$ is the angle between $\boldsymbol{v}$ and the horizontal plane, which can be computed by the arcsine of $v_z$ divided by $V$. Note that $v_z / V = \boldsymbol{e}_3^\top \boldsymbol{v} / \|\boldsymbol{v}\| = \boldsymbol{e}_3^\top \boldsymbol{r}_1$. We have

$$\gamma = -\arcsin(v_z / V) = -\arcsin(\boldsymbol{e}_3^\top \boldsymbol{r}_1) \tag{11}$$

where the negative sign $-$ is because $v_z$ points downwards in the NED frame, while $\gamma$ is positive upwards.

For fixed-wing vehicles, shown in Fig.2, the drag $D$ points to the opposite direction of $\boldsymbol{v}$. Under small angle-of-attack (AoA) conditions, the thrust $F_T$ aligns with the direction of $\boldsymbol{v}$. As for the lift, $L$ is always perpendicular $\boldsymbol{v}$. Through banking and coordinated turning, $L$ can be decomposed into two components for climbing and turning. Thus, except for gravity, all the other forces (i.e., $L$, $D$, $F_T$, and their resultant force is recorded as $\boldsymbol{F}_{n_g}$) can be conveniently projected into the flight path frame. One has $m\boldsymbol{a} = mg\boldsymbol{e}_3 + \boldsymbol{F}_{n_g}$. To facilitate the derivation of the flat mappings for controls, let $\boldsymbol{n}_g = \boldsymbol{F}_{n_g} / (mg) = \boldsymbol{a} / g - \boldsymbol{e}_3$ as the load vector in the inertial frame. Define the tangential, horizontal, and vertical load factors $n_x$, $n_y$, $n_z$ as the auxiliary controls, where $n_x = (F_T - D)/(mg)$, $n_y = L\sin\phi_b/(mg)$ and $n_z = L\cos\phi_b/(mg)$, and the practical controls $F_T$, $L$, and $\phi_b$ can be calculated by

$$F_T = n_x mg + D,\ L = mg\sqrt{n_y^2 + n_z^2},\ \phi_b = \text{acrtan}(n_y/n_z) \tag{12}$$

It can be found that $n_x, n_y, n_z$ are precisely the projections of $\boldsymbol{n}_g$ onto the flight path frame axes. Thus, the flat mappings of controls $\mathcal{F}_u:(\boldsymbol{p},\dot{\boldsymbol{p}},\ddot{\boldsymbol{p}})\to\boldsymbol{u}$ can be obtained as follows.

1) **Tangential load**. $n_x$ is the projection of $\boldsymbol{n}_g$ on $\boldsymbol{r}_1$

$$n_x = \boldsymbol{n}_g^\top\boldsymbol{r}_1 = \boldsymbol{n}_g^\top\boldsymbol{v}/\|\boldsymbol{v}\| \tag{13}$$

2) **Horizontal load**. $n_y$ is the projection of $\boldsymbol{n}_g$ on $\boldsymbol{r}_2$ as

$$n_y = \boldsymbol{n}_g^\top\boldsymbol{r}_2 = \boldsymbol{n}_g^\top(\boldsymbol{e}_3\times\boldsymbol{v})/\|\boldsymbol{e}_3\times\boldsymbol{v}\| \tag{14}$$

3) **Vertical load**. In Fig.2, a positive $n_z$ will drive the UAV to climb upwards, while $\boldsymbol{r}_3$ is downward for NED-frame. Thus, $n_z$ equals to the negative projection of $\boldsymbol{n}_g$ on $\boldsymbol{r}_3$ as

$$n_z = -\boldsymbol{n}_g^\top\boldsymbol{r}_3 = -\boldsymbol{n}_g^\top[\boldsymbol{v}\times(\boldsymbol{e}_3\times\boldsymbol{v})]/\|\boldsymbol{v}\times(\boldsymbol{e}_3\times\boldsymbol{v})\| \tag{15}$$

Finally, one can transcribe the nonlinear dynamics (1) as a differential flat integrator system through $\mathcal{F}_x$, $\mathcal{F}_u$ in (8)-(15).

$$\begin{aligned}\dot{\boldsymbol{p}}(t) &= \boldsymbol{v}(t)\\ \dot{\boldsymbol{v}}(t) &= \boldsymbol{a}(t)\end{aligned} \tag{16}$$

**Remark 3**. Equations (8)-(15) have singularities when $\|\boldsymbol{v}\| = 0$ or $\|\boldsymbol{e}_3\times\boldsymbol{v}\| = 0$. It is because, when the speed is zero or the flight direction is perpendicular to the horizontal plane, (1) is undefined. However, due to the practice minimum speed and the flight path angle constraints, the singularity of $\mathcal{F}_x$ and $\mathcal{F}_u$ will not happen in general cases, i.e., $\mathcal{F}_x$ and $\mathcal{F}_u$ can describe the bijection relationship from $(\boldsymbol{p},\dot{\boldsymbol{p}},\ddot{\boldsymbol{p}})$ to $(\boldsymbol{x},\boldsymbol{u})$ except for singular points. Thus, the inverse mappings, recorded as $\mathcal{F}_x^{-1}:\boldsymbol{x}\to(\boldsymbol{p},\dot{\boldsymbol{p}})$ and $\mathcal{F}_u^{-1}:(\boldsymbol{x},\boldsymbol{u})\to\ddot{\boldsymbol{p}}$, can also be obtained as

$$\begin{aligned}\begin{bmatrix}\boldsymbol{p}\\ \dot{\boldsymbol{p}}\end{bmatrix} &= \mathcal{F}_x^{-1}(\boldsymbol{x}) = [x,y,z,V\cos\gamma\cos\chi,V\cos\gamma\sin\chi,-V\sin\gamma]^\top\\ \ddot{\boldsymbol{p}} = \mathcal{F}_u^{-1}(\boldsymbol{x},\boldsymbol{u}) &= g\begin{bmatrix}\cos\gamma\cos\chi & -\sin\chi & -\cos\chi\sin\gamma\\ \cos\gamma\sin\chi & \cos\chi & -\sin\chi\sin\gamma\\ -\sin\gamma & 0 & -\cos\gamma\end{bmatrix}\begin{bmatrix}n_x\\ n_y\\ n_z\end{bmatrix}\\ &\quad+\begin{bmatrix}0 & 0 & g\end{bmatrix}^\top\end{aligned} \tag{17}$$

Therefore, based on the above dynamic transcription, one can transcribe **Problem 1** into solving the flat outputs $\boldsymbol{p}$ and their derivatives on linear state space instead of calculating the original states $\boldsymbol{x}$ and $\boldsymbol{u}$ on nonlinear dynamics, which brings convenience in subsequent fast trajectory planning. Once $\boldsymbol{p}$ is solved, $\boldsymbol{x}$ and $\boldsymbol{u}$ can be accurately determined by the flat mappings $\mathcal{F}_x$ and $\mathcal{F}_u$.

*B. Trajectory Polynomial Parameterization*

For the convenience of numerical trajectory optimization, owing to the approximation ability and the derivation simplicity of polynomials, the trajectory $\boldsymbol{p}(t)$ is parameterized by $N$ pieces of continuous polynomial splines as

$$\boldsymbol{p}(t) = \boldsymbol{C}_i^\top\boldsymbol{b}(t-t_{i-1}),\ t\in[t_{i-1},t_i],\ i=1,\ldots N \tag{18}$$

where $\boldsymbol{b}(t) = [1,t,t^2,\ldots,t^{s-1}]^\top\in\mathbb{R}^s$ is the polynomial basis, and $s$ denotes the order number of polynomials; $\boldsymbol{C}_i\in\mathbb{R}^{s\times3}$ is the coefficient matrix; $N$ is the number of segments; $t_{i-1}$ and $t_i$ are the start and final times of the segment $i$, respectively. The time intervals of each segment are recorded as $T_i = t_i - t_{i-1}$.

**Remark 4**. **Problem 1** requires at most the 2nd-order derivative $\ddot{\boldsymbol{p}}$ to obtain all the states and controls. In addition, Sec.IV.A introduces the 3rd-order derivative $\dddot{\boldsymbol{p}}$ additionally to smooth the trajectory. Therefore, refer to [20], we select $s=6$ to parameterize the trajectories of fixed-wing UAVs.

In (18), the parameterized variables are $T_i$ and $\boldsymbol{C}_i$, involving total $N(3s+1)$ parameters. Considering the practical dynamics constraints, these variables are not independent. To enhance the efficiency of trajectory optimization, these variables need to be decoupled and dimensionally reduced.

We first normalize and uniform the time variables to reduce temporal dimensions. Define $\tau = (t-t_{i-1})/T_i\in[0,1]$ as the normalized time. Equation (18) can be rewritten in a temporal-normalized form as

$$\boldsymbol{p}(t) = \bar{\boldsymbol{C}}_i^\top\boldsymbol{b}(\tau),\ \tau\in[0,1],\ i=1,\ldots N \tag{19}$$

where $\bar{\boldsymbol{C}}_i\in\mathbb{R}^{s\times3}$ is the temporal-normalized coefficient matrix. Then, let each segment have a uniform time duration, i.e., $T_i = T/N$ for $\forall i=1,\ldots,N$, in which $T = t_f - t_0 = \sum_{i=1}^N T_i$ is the total flight duration. This normalization and uniformization reduces the temporal variables of the polynomial spline to a single variable $T$. By comparing (18) and (19), the relationship between $\boldsymbol{C}_i$ and $\bar{\boldsymbol{C}}_i$ can be expressed as $\boldsymbol{C}_i = \boldsymbol{\theta}(T)\cdot\bar{\boldsymbol{C}}_i$, where $\boldsymbol{\theta}(T) = \text{diag}\{1,\frac{N}{T},\frac{N2}{T^2},\ldots,\frac{N^{s-1}}{T^{s-1}}\}\in\mathbb{R}^{s\times s}$. After the parameterization by (19), the derivatives of flat outputs $\boldsymbol{p}$ can be given as

$$\dot{\boldsymbol{p}} = \frac{N}{T}\bar{\boldsymbol{C}}_i^\top\frac{d\boldsymbol{b}}{d\tau},\ \ddot{\boldsymbol{p}} = \frac{N^2}{T^2}\bar{\boldsymbol{C}}_i^\top\frac{d^2\boldsymbol{b}}{d\tau^2},\ \ldots,\ \boldsymbol{p}^{(n)} = \frac{N^n}{T^n}\bar{\boldsymbol{C}}_i^\top\frac{d^n\boldsymbol{b}}{d\tau^n} \tag{20}$$

in which $\boldsymbol{p}^{(n)}$ represents the $n^{\text{th}}$-order derivatives of $\boldsymbol{p}$ with respect to $t$. For simplicity, in what follows, the derivatives of $\boldsymbol{b}$ over $\tau$ (e.g., $\frac{d\boldsymbol{b}}{d\tau}$, $\frac{d^2\boldsymbol{b}}{d\tau^2}$, and $\frac{d^n\boldsymbol{b}}{d\tau^n}$) are also abbreviated as $\dot{\boldsymbol{b}}$, $\ddot{\boldsymbol{b}}$,..., $\boldsymbol{b}^{(n)}$, etc.

After that, we reduce the dimensions of the polynomial coefficients. Considering the terminal constraints, (4) can be rewritten using flat outputs $\boldsymbol{p}$ as

$$\begin{cases}\boldsymbol{p}(t_0) = \bar{\boldsymbol{C}}_1^\top\boldsymbol{b}(0) = \boldsymbol{p}_0\\ \dot{\boldsymbol{p}}(t_0) = N\bar{\boldsymbol{C}}_1^\top\dot{\boldsymbol{b}}(0)/T = \boldsymbol{v}_0\\ \ddot{\boldsymbol{p}}(t_0) = N^2\bar{\boldsymbol{C}}_1^\top\ddot{\boldsymbol{b}}(0)/T^2 = \boldsymbol{a}_0\end{cases},\quad \begin{cases}\boldsymbol{p}(t_f) = \bar{\boldsymbol{C}}_N^\top\boldsymbol{b}(1) = \boldsymbol{p}_f\\ \dot{\boldsymbol{p}}(t_f) = N\bar{\boldsymbol{C}}_N^\top\dot{\boldsymbol{b}}(1)/T = \boldsymbol{v}_f\\ \ddot{\boldsymbol{p}}(t_f) = N^2\bar{\boldsymbol{C}}_N^\top\ddot{\boldsymbol{b}}(1)/T^2 = \boldsymbol{a}_f\end{cases} \tag{21}$$

where $\boldsymbol{p}_0,\boldsymbol{v}_0,\boldsymbol{a}_0,\boldsymbol{p}_f,\boldsymbol{v}_f,\boldsymbol{a}_f$ can be obtained from the terminal states and controls given in (4) and $\mathcal{F}_x^{-1}$, $\mathcal{F}_u^{-1}$ in (17). Besides, considering the smoothness and continuity of the polynomial spline, at the intermediate time $t_i = iT/N$ of any two adjacent segments, the flat outputs $\boldsymbol{p}$ at $t_i$ yield

$$\begin{cases}\boldsymbol{p}(t_i) = \bar{\boldsymbol{C}}_i^\top\boldsymbol{b}(1) = \bar{\boldsymbol{C}}_{i+1}^\top\boldsymbol{b}(0) = \boldsymbol{p}_i\\ \dot{\boldsymbol{p}}(t_i) = N\bar{\boldsymbol{C}}_i^\top\dot{\boldsymbol{b}}(1)/T = N\bar{\boldsymbol{C}}_{i+1}^\top\dot{\boldsymbol{b}}(0)/T\\ \quad\vdots\\ \boldsymbol{p}^{(4)}(t_i) = N^4\bar{\boldsymbol{C}}_i^\top\boldsymbol{b}^{(4)}(1)/T^4 = N^4\bar{\boldsymbol{C}}_{i+1}^\top\boldsymbol{b}^{(4)}(0)/T^4\end{cases} \tag{22}$$

in which $\boldsymbol{p}_i$ is the intermediate waypoints of two adjacent segments. (21)-(22) can be rewritten in matrix form as

$$\underline{\boldsymbol{B}}_0\bar{\boldsymbol{C}}_1 = \boldsymbol{D}_0,\ \bar{\boldsymbol{B}}_N\bar{\boldsymbol{C}}_N = \boldsymbol{D}_0,\ \text{and}\ \bar{\boldsymbol{B}}_i\bar{\boldsymbol{C}}_i + \underline{\boldsymbol{B}}_i\bar{\boldsymbol{C}}_{i+1} = \boldsymbol{D}_i, i=1,\ldots,N-1 \tag{23}$$

in which

$$\underline{\boldsymbol{B}}_0 = \left[\boldsymbol{b}(0), \dot{\boldsymbol{b}}(0), \ddot{\boldsymbol{b}}(0)\right]^\top \in \mathbb{R}^{3\times6},\ \boldsymbol{D}_0 = \left[\boldsymbol{p}_0, \tfrac{T}{N}\boldsymbol{v}_0, \tfrac{T^2}{N^2}\boldsymbol{a}_0\right]^\top \in \mathbb{R}^{3\times3}$$
$$\bar{\boldsymbol{B}}_N = \left[\boldsymbol{b}(1), \dot{\boldsymbol{b}}(1), \ddot{\boldsymbol{b}}(1)\right]^\top \in \mathbb{R}^{3\times6},\ \boldsymbol{D}_N = \left[\boldsymbol{p}_f, \tfrac{T}{N}\boldsymbol{v}_f, \tfrac{T^2}{N^2}\boldsymbol{a}_f\right]^\top \in \mathbb{R}^{3\times3} \quad (24)$$
$$\underline{\boldsymbol{B}}_i = \left[\boldsymbol{0}_{6\times1}, -\boldsymbol{b}(0), -\dot{\boldsymbol{b}}(0), \ldots, -\boldsymbol{b}^{(4)}(0)\right]^\top \in \mathbb{R}^{6\times6}$$
$$\bar{\boldsymbol{B}}_i = \left[\boldsymbol{b}(1), \boldsymbol{b}(1), \dot{\boldsymbol{b}}(1), \ldots, \boldsymbol{b}^{(4)}(1)\right]^\top \in \mathbb{R}^{6\times6},\ \boldsymbol{D}_i = \left[\boldsymbol{p}_i, \boldsymbol{0}_{3\times5}\right]^\top \in \mathbb{R}^{6\times3}$$

Furthermore, (23) can be summarized as

$$\mathcal{B} \cdot \bar{\mathcal{C}}(\mathcal{P}, T) = \mathcal{D}(\mathcal{P}, T) \quad (25)$$

where $\mathcal{P} = [\boldsymbol{p}_1, \boldsymbol{p}_2, \ldots, \boldsymbol{p}_i, \ldots, \boldsymbol{p}_{N-1}]^\top \in \mathbb{R}^{(N-1)\times3}$ is the arrangement of intermediate waypoints; $\bar{\mathcal{C}} = \mathrm{col}\{\bar{\mathcal{C}}_1, \bar{\mathcal{C}}_2, \ldots, \bar{\mathcal{C}}_i, \ldots, \bar{\mathcal{C}}_N\} \in \mathbb{R}^{6N\times3}$ represents the column arrangement of polynomial coefficients; and $\mathcal{B}$, $\mathcal{D}$ are detailed as

$$\mathcal{B} = \begin{bmatrix} \underline{\boldsymbol{B}}_0 & \boldsymbol{0} & \boldsymbol{0} & \cdots & \boldsymbol{0} & \boldsymbol{0} & \cdots & \boldsymbol{0} & \boldsymbol{0} \\ \bar{\boldsymbol{B}}_1 & \underline{\boldsymbol{B}}_1 & \boldsymbol{0} & \cdots & \boldsymbol{0} & \boldsymbol{0} & \cdots & \boldsymbol{0} & \boldsymbol{0} \\ \boldsymbol{0} & \bar{\boldsymbol{B}}_2 & \underline{\boldsymbol{B}}_2 & \cdots & \boldsymbol{0} & \boldsymbol{0} & \cdots & \boldsymbol{0} & \boldsymbol{0} \\ \vdots & \vdots & \vdots & \ddots & \vdots & \vdots & \ddots & \vdots & \vdots \\ \boldsymbol{0} & \boldsymbol{0} & \boldsymbol{0} & \cdots & \bar{\boldsymbol{B}}_i & \underline{\boldsymbol{B}}_i & \cdots & \boldsymbol{0} & \boldsymbol{0} \\ \vdots & \vdots & \vdots & \ddots & \vdots & \vdots & \ddots & \vdots & \vdots \\ \boldsymbol{0} & \boldsymbol{0} & \boldsymbol{0} & \cdots & \boldsymbol{0} & \boldsymbol{0} & \cdots & \bar{\boldsymbol{B}}_{N-1} & \underline{\boldsymbol{B}}_{N-1} \\ \boldsymbol{0} & \boldsymbol{0} & \boldsymbol{0} & \cdots & \boldsymbol{0} & \boldsymbol{0} & \cdots & \boldsymbol{0} & \bar{\boldsymbol{B}}_N \end{bmatrix} \in \mathbb{R}^{6N\times6N} \quad (26)$$
$$\mathcal{D} = \mathrm{col}\{\boldsymbol{D}_0, \boldsymbol{D}_1, \ldots, \boldsymbol{D}_i, \ldots, \boldsymbol{D}_N\} \in \mathbb{R}^{6N\times3}$$

Note that $\mathcal{B}$ is a non-singular band matrix. Then, the relationship between the $\bar{\mathcal{C}}$ with $\mathcal{P}$, $T$ can be given as

$$\bar{\mathcal{C}}(\mathcal{P}, T) = \mathcal{B}^{-1} \cdot \mathcal{D}(\mathcal{P}, T) \quad (27)$$

which indicates that we can obtain the polynomial coefficients $\bar{\mathcal{C}}$ by computing the intermediate waypoints $\mathcal{P}$ and the total flight duration $T$. Thus, the polynomial spline parameters are decoupled and dimensionality reduced:

1) **Decoupling**. $\mathcal{P}$ denotes the intermediate waypoints, and $T$ is the total flight duration. $\mathcal{P}$ and $T$ are decoupled spatial and temporal variables, respectively. It simplifies subsequent constraint processing and convenience trajectory optimization.
2) **Dimensionality Reduction**. Initially, the polynomials are determined by $T_i$ and $\boldsymbol{C}_i$, involving $N(3s+1)$ variables. Through (19)-(27), the variables are tightly reduced to $\mathcal{P}$ and $T$, leaving only $3(N-1)+1$ ones, which improves the efficiency of trajectory optimization.

**Remark 5**. Owing to the temporal normalization and uniformization in (19), $\mathcal{B}$ only contains $\boldsymbol{b}$ and its derivatives at $\tau = 0$ and $\tau = 1$. Therefore, $\mathcal{B}$ is already constant. $\mathcal{B}^{-1}$ in (27) can be pre-calculated and stored to avoid repeated matrix inverse operations and save computational time.

Finally, combining the differential flatness of fixed-wing UAVs in Sec.III.A, we obtain a class of trajectories compactly parameterized by $\mathcal{P}$ and $T$ shown as

$$\mathrm{Traj}_{\mathrm{FW}} = \big\{\boldsymbol{p}(t) : [0, T] \mid \bar{\mathcal{C}} = \bar{\mathcal{C}}(\mathcal{P}, T), \mathcal{P} \in \mathbb{R}^{3(N-1)}, T \in \mathbb{R}_+ \\ \text{and } \boldsymbol{x} = \mathcal{F}_x(\boldsymbol{p}, \dot{\boldsymbol{p}}),\ \boldsymbol{u} = \mathcal{F}_u(\boldsymbol{p}, \dot{\boldsymbol{p}}, \ddot{\boldsymbol{p}})\big\} \quad (28)$$

Utilizing the trajectory representation model $\mathrm{Traj}_{\mathrm{FW}}$ in (28), one can transform the **Problem 1** into

$$\textbf{Problem 2:} \quad \min_{\mathcal{P}, T}\ \mathcal{J} = \mathcal{Q}(T) + \lambda \int_{t_0}^{t_f} \mathcal{G}\left[\boldsymbol{x}(\mathcal{P}, T), \boldsymbol{u}(\mathcal{P}, T)\right] dt$$
$$\text{s.t.}\ \ \boldsymbol{x}_{\min} \le \boldsymbol{x}(\mathcal{P}, T) \le \boldsymbol{x}_{\max},\ \boldsymbol{u}_{\min} \le \boldsymbol{u}(\mathcal{P}, T) \le \boldsymbol{u}_{\max} \quad (29)$$
$$\left\|\boldsymbol{G} \cdot \boldsymbol{x}(\mathcal{P}, T) - \boldsymbol{p}_{obs,j}\right\| \ge R_{obs,j} + R_{safe},\ \forall j \in \mathcal{O}_{obs}$$

**Remark 6**. Leveraging $\mathrm{Traj}_{\mathrm{FW}}$, the original states and controls $\boldsymbol{x}, \boldsymbol{u}$ are replaced by flat outputs $\boldsymbol{p}$ through flat mappings $\mathcal{F}_x, \mathcal{F}_u$ in (8)-(15), and further parameterized as polynomials determined with the intermediate waypoints $\mathcal{P}$ and the flight duration $T$ by (19)-(27). Since all the above transformations involve direct variable substitutions via equations, the transcription process from the original problem to **Problem 2** preserves equivalence. Rather than directly optimizing complex nonlinear dynamics, **Problem 2** eliminates all the equality dynamics and terminal constraints of the original problem, avoiding the heavy computational burdens. Moreover, by focusing on solving the decoupled and lower-dimensional variables $\mathcal{P}$, $T$ in flat spaces, **Problem 2** significantly reduces the complexity of trajectory optimization. Once $\mathcal{P}$ and $T$ are solved, the polynomial coefficients $\bar{\mathcal{C}}$ can be efficiently computed to determine the trajectory's shape. Subsequently, the original state spaces $\boldsymbol{x}, \boldsymbol{u}$ can be recovered indirectly by $\mathcal{F}_x, \mathcal{F}_u$, yielding accurate results of trajectory states. Therefore, **Problem 2** demonstrates the potential for efficient trajectory resolution.

However, since **Problem 2** remains inequality constraints, it is still computationally intensive. The following section will further formulate **Problem 2** into a lightweight, unconstrained, gradient-analytical optimization to enable more efficient trajectory computation.

## IV. Lightweight Unconstrained Trajectory Optimization with Analytical Gradients

This section proposes a lightweight trajectory optimization framework for efficient trajectory generation of fixed-wing UAVs. By designing penalties to satisfy inequality constraints, we transcribe **Problem 2** into an unconstrained optimization. Then, the analytical gradients are derived to accelerate optimization iterations. Finally, the detailed procedure of trajectory generation for fixed-wing UAVs is presented. The details are as follows.

### *A. Unconstrained Trajectory Optimization*

In this subsection, **Problem 2** is further transformed into an unconstrained optimization problem as follows

$$\textbf{Problem 3}: \quad \min_{\mathcal{P}, T}\ \mathcal{J}(\mathcal{P}, T) = \mathcal{Q}(T) + \sum \lambda_\sigma \mathcal{I}_\sigma(\bar{\mathcal{C}}, T) \quad (30)$$

in which, $\mathcal{Q}$ is the time-related cost; $\mathcal{I}_\sigma$ denotes the integral costs with $\sigma \in \{e, V, \gamma, n_x, n_y, n_z, obs\}$, covering control effort $\mathcal{I}_e$, and penalties on obstacle avoidance $\mathcal{I}_{obs}$, states $\mathcal{I}_V$, $\mathcal{I}_\gamma$, and loads $\mathcal{I}_{n_x}$, $\mathcal{I}_{n_y}$, $\mathcal{I}_{n_z}$; and each $\lambda_\sigma$ is a corresponding weight. The detailed design of these cost functions is provided below.

1) Time-related Cost

Trajectory optimization usually aims to minimize the flight duration and find the fastest trajectory to reach the target position. In this case, $\mathcal{Q}$ is given as

$$\mathcal{Q}(T) = T \tag{31}$$

2) Integral Costs

The integral costs $\mathcal{I}_\sigma$ are designed to impose constraints on the states and control inputs throughout the flight. Since the trajectory is parameterized in a piecewise manner, $\mathcal{I}_\sigma$ can be calculated as the sum of the cost over each segment, recorded as $\mathcal{I}_{\sigma,i}$. Use the trapezoidal integral method to calculate $\mathcal{I}_\sigma$ numerically as

$$\mathcal{I}_\sigma(\bar{\mathcal{C}},T) = \sum_{i=1}^{N} \mathcal{I}_{\sigma,i} = \sum_{i=1}^{N} \int_0^{T/N} \mathcal{G}_\sigma(\bar{\boldsymbol{C}}_i,t)dt = \sum_{i=1}^{N} \frac{T}{\kappa N} \sum_{k=0}^{\kappa} \omega_k \mathcal{G}_\sigma(\bar{\boldsymbol{C}}_i,t_k) \tag{32}$$

in which, $t_k = \frac{kT}{N\kappa}$ denotes the sample time for trapezoidal integral calculation with $k = 0,1,2,\ldots,\kappa$, and $\kappa$ is the number of samples. $\omega_k \in \boldsymbol{\omega} = [\frac{1}{2},1,\ldots,1,\frac{1}{2}]^\top \in \mathbb{R}^{\kappa+1}$ refers to the trapezoidal integration weight. For $k = 0$ or $k = \kappa$, $\omega_k = \frac{1}{2}$; otherwise, $\omega_k = 1$. $\mathcal{G}_\sigma$ with $\sigma \in \{e,V,\gamma,n_x,n_y,n_z,obs\}$ represents the integrand to be designed as follows.

a) **Control effort cost**. The 3rd-order derivative $\dddot{\boldsymbol{p}}$, denoted as the jerk vector $\boldsymbol{j} \in \mathbb{R}^3$, is introduced to minimize the control efforts and smooth the trajectory. Thus, $\mathcal{G}_e$ is given by

$$\mathcal{G}_e(\bar{\boldsymbol{C}}_i,t) = \boldsymbol{j}^\top \boldsymbol{j} \tag{33}$$

b) **Obstacle avoidance penalty**. As for the obstacle avoidance constraint in (3), we design the distance field as the penalty to drive the UAV to stay away from the obstacle as

$$\phi_{obs,j} = 1 - \left[ \frac{\|\bar{\boldsymbol{H}} \cdot \boldsymbol{p}(t) - \boldsymbol{p}_{obs,j}\|}{\zeta_{obs} \cdot (R_{obs,j} + R_{safe})} \right]^2, \ j \in \mathcal{O}_{obs} \tag{34}$$

in which, $\bar{\boldsymbol{H}} = [\boldsymbol{I}_{2\times2}, \boldsymbol{0}_{2\times1}] \in \mathbb{R}^{2\times3}$ is the selection matrix of *x*, *y* coordinates, and $\|\bar{\boldsymbol{H}} \cdot \boldsymbol{p}(t) - \boldsymbol{p}_{obs,j}\|$ represents the distance to the obstacle *j*; $\zeta_{obs} > 1$ is the threshold factor to shrink the constraints. When the UAV is close to the obstacle, $\phi_{obs,j}$ will be positive. Thus, we design the following penalty to decrease $\phi_{obs,j}$ for collision avoidance.

$$\mathcal{G}_{obs}(\bar{\boldsymbol{C}}_i,t) = \sum_{j\in\mathcal{O}_{obs}} \mathcal{G}_{obs,j}(\bar{\boldsymbol{C}}_i,t) = \sum_{j\in\mathcal{O}_{obs}} \left\{\max\left[\phi_{obs,j}(\bar{\boldsymbol{C}}_i,t),0\right]\right\}^{\varpi} \tag{35}$$

where $\varpi \in \mathbb{N}_+$ is the power exponent.

**Remark 7**. To enhance the numerical performance of the penalty across different scenarios, the magnitude of (34) has been normalized by the constraint boundary $R_{obs,j} + R_{safe}$. In addition, the threshold factor $\zeta_{obs}$ is introduced to shrink the constraints and improve the safety margin of the trajectory. Such strategies are also utilized in the subsequent penalty design to ensure feasibility for effective constraint processing.

c) **Flight speed penalty**. $V$ is limited by $V_{\min} \le V \le V_{\max}$. The constraint can be rewritten as $-V_h \le V - V_c \le V_h$, where $V_c = (V_{\max} + V_{\min})/2$ and $V_h = (V_{\max} - V_{\min})/2$. Then, shrink the boundary as $-\zeta_V V_h \le V - V_c \le \zeta_V V_h$ with the threshold factor $\zeta_V < 1$, and the penalty of flight speed constraint is given as

$$\mathcal{G}_V(\bar{\boldsymbol{C}}_i,t) = \left\{\max\left[\phi_V(\bar{\boldsymbol{C}}_i,t),0\right]\right\}^{\varpi}, \ \phi_V = \left(\frac{V - V_c}{\zeta_V V_h}\right)^2 - 1 \tag{36}$$

d) **Flight path angle penalty**. $\gamma$ is constrained by $\gamma_{\min} \le \gamma \le \gamma_{\max}$. Note that sine is monotonically increasing on $[-\frac{\pi}{2},\frac{\pi}{2}]$. Therefore, the constraint can be transformed as $\sin\gamma_{\min} \le \sin\gamma \le \sin\gamma_{\max}$. Let $\Gamma_c = (\sin\gamma_{\max} + \sin\gamma_{\min})/2$, $\Gamma_h = (\sin\gamma_{\max} - \sin\gamma_{\min})/2$, we have $-\Gamma_h \le \sin\gamma - \Gamma_c \le \Gamma_h$ with $\sin\gamma = -\boldsymbol{e}_3^\top \boldsymbol{r}_1$. Then, similarly to (36), the penalty can be designed as

$$\mathcal{G}_\gamma(\bar{\boldsymbol{C}}_i,t) = \left\{\max\left[\phi_\gamma(\bar{\boldsymbol{C}}_i,t),0\right]\right\}^{\varpi}, \ \phi_\gamma = \left(\frac{-\boldsymbol{e}_3^\top \boldsymbol{r}_1 - \Gamma_c}{\zeta_\gamma \Gamma_h}\right)^2 - 1 \tag{37}$$

in which $\zeta_\gamma < 1$ is the corresponding threshold factor.

e) **Penalties for loads**. The control load factors are bounded by limited maneuverability, i.e., $n_{x,\min} \le n_x \le n_{x,\max}$, $n_{y,\min} \le n_y \le n_{y,\max}$, and $n_{z,\min} \le n_z \le n_{z,\max}$. For simplicity of expression, record $n_x,n_y,n_z$ as the symbol $n_l$ with $l \in \{x,y,z\}$, and thus the constraints can be recorded as $n_{l,\min} \le n_l \le n_{l,\max}$. Let $n_{l,c} = (n_{l,\min} + n_{l,\max})/2$, $n_{l,h} = (n_{l,\max} - n_{l,\min})/2$, and the constraints can be further transcribed as $-n_{l,h} \le n_l - n_{l,c} \le n_{l,h}$. Then, the penalty terms of load factors are designed as

$$\mathcal{G}_{nl}(\bar{\boldsymbol{C}}_i,t) = \left\{\max\left[\phi_{nl}(\bar{\boldsymbol{C}}_i,t),0\right]\right\}^{\varpi}, \ \phi_{nl} = \left(\frac{n_l - n_{l,c}}{\zeta_{nl} n_{l,h}}\right)^2 - 1 \tag{38}$$

where $l \in \{x,y,z\}$, and $\zeta_{nl}$ means the threshold factor.

3) Elimination of Implicit Time Constraint

Notice that **Problem 2** has an implicit constraint, i.e., the flight duration $T$ must be positive to avoid singularity during the optimization of minimizing flight time. To address this issue, we replace the optimization variable $T$ by $\mathcal{T}$ as

$$T = \exp(\mathcal{T}), \ \mathcal{T} = \ln(T) \tag{39}$$

in which $T > 0$ is ensured strictly for any $\mathcal{T} \in \mathbb{R}$.

### *B. Analytical Gradients Derivation*

Typical nonlinear optimization approaches for complex dynamics often rely on finite difference methods to obtain gradients for iteration, resulting in significant computational burdens. In contrast, by utilizing the trajectory representation $\text{Traj}_{\text{FW}}$ and the design of penalties, this paper transcribes the original problem into lightweight **Problem 3**, which has a simplified mathematical form and allows for straightforward analytical differentiation. Therefore, this subsection derives the analytical gradients to speed up the trajectory optimization process significantly.

1) Spatial-temporal Derivatives of $\text{Traj}_{\text{FW}}$

Since the trajectory optimization has been reformulated to determine $\bar{\mathcal{C}}$ indirectly by optimizing $\boldsymbol{\mathcal{P}}$ and $T$, we first deduct the analytical spatial-temporal derivative relationships of the trajectory representation $\text{Traj}_{\text{FW}}$ with respect to these variables. Consider the functional objective $\mathcal{J}$ related to $\boldsymbol{\mathcal{P}}$, $T$ (optimization variables) as

$$\mathcal{J}(\boldsymbol{\mathcal{P}},T) = \mathcal{L}\left[\boldsymbol{\Theta}(T) \cdot \bar{\mathcal{C}}(\boldsymbol{\mathcal{P}},T), \ T\right] \tag{40}$$

where $\mathcal{L}$ on the right side represents the functional cost related to $\bar{\mathcal{C}}$ and $T$ (i.e. the costs designed in Sec.IV.A). $\boldsymbol{\Theta}(T) = \text{diag}\{\boldsymbol{\theta}(T),\ldots,\boldsymbol{\theta}(T)\} \in \mathbb{R}^{6N\times6N}$ denotes the arrangement matrix of $\boldsymbol{\theta}(T)$ obtained from the temporal normalization of the polynomials in (19). Then, the derivative relationships between $\frac{\partial \mathcal{J}}{\partial \mathcal{P}}$, $\frac{\partial \mathcal{J}}{\partial T}$ with $\frac{\partial \mathcal{L}}{\partial \bar{\mathcal{C}}}$, $\frac{\partial \mathcal{L}}{\partial T}$ are calculated as follows.

a) To calculate $\frac{\partial \mathcal{J}}{\partial \mathcal{P}}$, differentiate (27) from both sides by $p_{i,j}$ with $i=1,\ldots,N-1$ and $j=1,2,3$

$$\frac{\partial \bar{\mathcal{C}}}{\partial p_{i,j}} = \mathcal{B}^{-1}\frac{\partial \mathcal{D}}{\partial p_{i,j}} \tag{41}$$

According to the chain derivative rule, one has

$$\begin{aligned}\frac{\partial \mathcal{J}}{\partial p_{i,j}} &= \mathrm{Tr}\left[\left(\frac{\partial \bar{\mathcal{C}}}{\partial p_{i,j}}\right)^{\top}\frac{\partial \mathcal{L}}{\partial \bar{\mathcal{C}}}\right] = \mathrm{Tr}\left[\left(\mathcal{B}^{-1}\frac{\partial \mathcal{D}}{\partial p_{i,j}}\right)^{\top}\frac{\partial \mathcal{L}}{\partial \bar{\mathcal{C}}}\right] \\ &= \mathrm{Tr}\left[\left(\frac{\partial \mathcal{D}}{\partial p_{i,j}}\right)^{\top}\left(\mathcal{B}^{-\top}\frac{\partial \mathcal{L}}{\partial \bar{\mathcal{C}}}\right)\right]\end{aligned} \tag{42}$$

where $\mathrm{Tr}(\cdot)$ denotes the trace of a matrix. Note that $p_{i,j}$ only occurs at $3(2i-1)+1^{\text{th}}$ row and $j^{\text{th}}$ column in (24)-(26), and thus $\frac{\partial \mathcal{D}}{\partial p_{i,j}}$ has one and only one nonzero element 1 at the same place. Let $\boldsymbol{G} = \mathcal{B}^{-\top}\cdot\frac{\partial \mathcal{L}}{\partial \bar{\mathcal{C}}} = \mathrm{col}\{\boldsymbol{G}_0,\boldsymbol{G}_1,\ldots,\boldsymbol{G}_{N-1},\boldsymbol{G}_N\} \in \mathbb{R}^{6N\times 3}$, where $\boldsymbol{G}_0,\boldsymbol{G}_N \in \mathbb{R}^{3\times 3}$, $\boldsymbol{G}_i \in \mathbb{R}^{6\times 3}$. Then, rewrite (42) as

$$\frac{\partial \mathcal{J}}{\partial \mathcal{P}} = \mathrm{col}\{\boldsymbol{\delta}^{\top}\boldsymbol{G}_1,\boldsymbol{\delta}^{\top}\boldsymbol{G}_2,\ldots,\boldsymbol{\delta}^{\top}\boldsymbol{G}_{N-1}\} \tag{43}$$

in which $\boldsymbol{\delta} = [1,0,0,\ldots,0]^{\top} \in \mathbb{R}^6$.

b) To obtain $\frac{\partial \mathcal{J}}{\partial T}$, differentiate (27) from both sides by $T$ as

$$\frac{\partial \bar{\mathcal{C}}}{\partial T} = \mathcal{B}^{-1}\frac{\partial \mathcal{D}}{\partial T} \tag{44}$$

Note that $T$ only occurs in the 1st, 2nd, $6N-1^{\text{th}}$, and $6N^{\text{th}}$ rows of $\mathcal{D}$ in (24)-(26). Thus, one has

$$\begin{aligned}\frac{\partial \mathcal{D}}{\partial T} &= \left[\frac{\partial \boldsymbol{D}_0}{\partial T},\mathbf{0},\ldots,\mathbf{0},\frac{\partial \boldsymbol{D}_N}{\partial T}\right]^{\top} \\ \frac{\partial \boldsymbol{D}_0}{\partial T} &= \left[\mathbf{0},\frac{\boldsymbol{v}_0}{N},\frac{2T\boldsymbol{a}_0}{N^2}\right]^{\top},\ \frac{\partial \boldsymbol{D}_N}{\partial T} = \left[\mathbf{0},\frac{\boldsymbol{v}_f}{N},\frac{2T\boldsymbol{a}_f}{N^2}\right]^{\top}\end{aligned} \tag{45}$$

Then, differentiating (40) from both sides by $T$ and substituting (45) leads to

$$\begin{aligned}\frac{\partial \mathcal{J}}{\partial T} &= \frac{\partial \mathcal{L}(\boldsymbol{\Theta}\cdot\bar{\mathcal{C}},T)}{\partial T} = \frac{\partial \mathcal{L}}{\partial T} + \mathrm{Tr}\left[\left(\frac{\partial \mathcal{L}}{\partial(\boldsymbol{\Theta}\cdot\bar{\mathcal{C}})}\right)^{\top}\cdot\frac{\partial(\boldsymbol{\Theta}\cdot\bar{\mathcal{C}})}{\partial T}\right] \\ &= \frac{\partial \mathcal{L}}{\partial T} + \mathrm{Tr}\left\{\left(\boldsymbol{\Theta}^{-1}\cdot\frac{\partial \mathcal{L}}{\partial \bar{\mathcal{C}}}\right)^{\top}\cdot\left[\boldsymbol{\Theta}\cdot\left(\mathcal{B}^{-1}\cdot\frac{\partial \mathcal{D}}{\partial T}\right)+\frac{\partial \boldsymbol{\Theta}}{\partial T}\cdot\bar{\mathcal{C}}\right]\right\}\end{aligned} \tag{46}$$

2) Derivatives of Time-related Cost

For $\mathcal{Q}$ in (31), it is independent with the coefficients $\bar{\mathcal{C}}$. Therefore, the derivatives of $\mathcal{Q}$ with respect to $\bar{\mathcal{C}}$ and $T$ are given as

$$\frac{\partial \mathcal{Q}}{\partial \bar{\mathcal{C}}} = \mathbf{0},\ \frac{\partial \mathcal{Q}}{\partial T} = 1 \tag{47}$$

3) Derivatives of Integral Costs

$\mathcal{I}_{\sigma}$ in (32) can be differentiated by $\bar{\mathcal{C}}$ and $T$ as

$$\frac{\partial \mathcal{I}_{\sigma}}{\partial \bar{\mathcal{C}}} = \mathrm{col}\left\{\frac{\partial \mathcal{I}_{\sigma,1}}{\partial \bar{\boldsymbol{C}}_1},\ldots,\frac{\partial \mathcal{I}_{\sigma,i}}{\partial \bar{\boldsymbol{C}}_i},\ldots,\frac{\partial \mathcal{I}_{\sigma,N}}{\partial \bar{\boldsymbol{C}}_N}\right\},\ \frac{\partial \mathcal{I}_{\sigma}}{\partial T} = \sum_{i=1}^{N}\frac{\partial \mathcal{I}_{\sigma,i}}{\partial T} \tag{48}$$

in which, $\frac{\partial \mathcal{I}_{\sigma,i}}{\partial \bar{\boldsymbol{C}}_i}$ and $\frac{\partial \mathcal{I}_{\sigma,i}}{\partial T}$ are

$$\begin{aligned}\frac{\partial \mathcal{I}_{\sigma,i}}{\partial \bar{\boldsymbol{C}}_i} &= \frac{T}{\kappa N}\sum_{k=0}^{\kappa}\omega_k\cdot\frac{\partial \mathcal{G}_{\sigma}}{\partial \bar{\boldsymbol{C}}_i}(t_k) \\ \frac{\partial \mathcal{I}_{\sigma,i}}{\partial T} &= \frac{\mathcal{I}_{\sigma,i}}{T} + \frac{T}{\kappa N}\sum_{k=0}^{\kappa}\omega_k\cdot\frac{k-1}{\kappa}\cdot\frac{\partial \mathcal{G}_{\sigma}}{\partial t}(t_k)\end{aligned} \tag{49}$$

where $\frac{\partial \mathcal{G}_{\sigma}}{\partial \bar{\boldsymbol{C}}_i}$, $\frac{\partial \mathcal{G}_{\sigma}}{\partial t}$ for $\sigma \in \{e,V,\gamma,n_x,n_y,n_z,obs\}$ are detailed as follows.

a) **Control effort cost**. Taking derivatives of (33) with respect to $\bar{\boldsymbol{C}}_i$ and $t$ leads to

$$\frac{\partial \mathcal{G}_e}{\partial \bar{\boldsymbol{C}}_i} = 2\frac{N^3}{T^3}\ddot{\boldsymbol{b}}\boldsymbol{j}^{\top},\ \frac{\partial \mathcal{G}_e}{\partial t} = 2\boldsymbol{s}^{\top}\boldsymbol{j} \tag{50}$$

in which $\boldsymbol{s} = \dot{\boldsymbol{j}} = N^4\bar{\boldsymbol{C}}_i^{\top}\boldsymbol{b}^{(4)}(\tau)/T^4$ is the snap vector.

Note that, since the penalties $\mathcal{G}_{\sigma}$ in (35)-(38) have similar power forms $\max(\cdot)^{\varpi}$, one has

$$\frac{\partial \mathcal{G}_{\sigma}}{\partial \phi_{\sigma}} = \varpi\cdot\left\{\max\left[\phi_{\sigma}(\bar{\boldsymbol{C}}_i,t),0\right]\right\}^{\varpi-1} \tag{51}$$

for each $\sigma \in \{V,\gamma,n_x,n_y,n_z,obs_j\}$. Thus, through the chain rule, only $\frac{\partial \phi_{\sigma}}{\partial \bar{\boldsymbol{C}}_i}$, $\frac{\partial \phi_{\sigma}}{\partial t}$ needs to be given as follows.

b) **Obstacle avoidance penalty**. Differentiating $\phi_{obs,j}$ gives

$$\frac{\partial \phi_{obs,j}}{\partial \bar{\boldsymbol{C}}_i} = \frac{2\boldsymbol{b}(\bar{\boldsymbol{H}}\cdot\boldsymbol{p}-\boldsymbol{p}_{obs,j})^{\top}\bar{\boldsymbol{H}}}{\left[\zeta_{obs}(R_{obs,j}+R_{safe})\right]^2},\frac{\partial \phi_{obs,j}}{\partial t} = \frac{2(\bar{\boldsymbol{H}}\cdot\boldsymbol{v})^{\top}(\bar{\boldsymbol{H}}\cdot\boldsymbol{p}-\boldsymbol{p}_{obs,j})}{\left[\zeta_{obs}(R_{obs,j}+R_{safe})\right]^2} \tag{52}$$

c) **Flight speed penalty**. Differentiating $\phi_V$ gives

$$\frac{\partial \phi_V}{\partial \bar{\boldsymbol{C}}_i} = \frac{1}{(\zeta_V V_h)^2}\frac{V-V_c}{V}\frac{N}{T}\dot{\boldsymbol{b}}\boldsymbol{v}^{\top},\ \frac{\partial \phi_V}{\partial t} = \frac{1}{(\zeta_V V_h)^2}\frac{V-V_c}{V}\boldsymbol{a}^{\top}\boldsymbol{v} \tag{53}$$

d) **Flight path angle penalty**. One can differentiate (37) over $\bar{\boldsymbol{C}}_i$ and $t$ as

$$\begin{aligned}&\frac{\partial \phi_{\gamma}}{\partial t} = -\frac{2(\sin\gamma-\Gamma_c)}{(\zeta_{\gamma}\Gamma_h)^2}\boldsymbol{e}_3^{\top}\frac{\partial \boldsymbol{r}_1}{\partial t},\ \frac{\partial \phi_{\gamma}}{\partial \bar{\boldsymbol{C}}_i} = \frac{2(\sin\gamma-\Gamma_c)}{(\zeta_{\gamma}\Gamma_h)^2}\frac{\partial(\boldsymbol{e}_3^{\top}\boldsymbol{r}_1)}{\partial \bar{\boldsymbol{C}}_i} \\ &\frac{\partial \boldsymbol{r}_1}{\partial t} = \boldsymbol{r}_1\times\frac{\boldsymbol{a}}{\|\boldsymbol{v}\|}\times\boldsymbol{r}_1,\ \frac{\partial(\boldsymbol{e}_3^{\top}\boldsymbol{r}_1)}{\partial \bar{\boldsymbol{C}}_i} = \frac{N}{T}\frac{\dot{\boldsymbol{b}}}{\|\boldsymbol{v}\|}\boldsymbol{e}_3^{\top}\left(\boldsymbol{I}-\frac{\boldsymbol{v}\boldsymbol{v}^{\top}}{\boldsymbol{v}^{\top}\boldsymbol{v}}\right)\end{aligned} \tag{54}$$

e) **Penalties for loads**. Differentiate $\phi_{nl}$ over $\bar{\boldsymbol{C}}_i$ and $t$ for different $l \in \{x,y,z\}$ leads to

$$\begin{aligned}\frac{\partial \phi_{nl}}{\partial t} &= 2\mu\frac{\mu\boldsymbol{n}_g^{\top}\boldsymbol{r}_{\rho}-n_{l,c}}{(\zeta_{nl}n_{l,h})^2}\left(\boldsymbol{n}_g^{\top}\frac{\partial \boldsymbol{r}_{\rho}}{\partial t}+\boldsymbol{r}_{\rho}^{\top}\frac{\partial \boldsymbol{n}_g}{\partial t}\right) \\ \frac{\partial \phi_{nl}}{\partial \bar{\boldsymbol{C}}_i} &= 2\mu\frac{\mu\boldsymbol{n}_g^{\top}\boldsymbol{r}_{\rho}-n_{l,c}}{(\zeta_{nl}n_{l,h})^2}\frac{\partial(\boldsymbol{n}_g^{\top}\boldsymbol{r}_{\rho})}{\partial \bar{\boldsymbol{C}}_i}\end{aligned} \tag{55}$$

where $\mu \in \{+1,+1,-1\}$ and $\rho \in \{1,2,3\}$ for each $l \in \{x,y,z\}$, respectively. ($\mu=-1$ is because $n_z$ equals to the negative projection of $\boldsymbol{n}_g$ on $\boldsymbol{r}_3$); $\frac{\partial \boldsymbol{n}_g}{\partial t} = \boldsymbol{j}/g$; and $\frac{\partial \boldsymbol{r}_{\rho}}{\partial t}$, $\frac{\partial(\boldsymbol{n}_g^{\top}\boldsymbol{r}_{\rho})}{\partial \bar{\boldsymbol{C}}_i}$ for different $\rho \in \{1,2,3\}$ are detailed as

$$\begin{aligned}&\frac{\partial \boldsymbol{r}_1}{\partial t} = \boldsymbol{r}_1\times\frac{\boldsymbol{a}}{\|\boldsymbol{v}\|}\times\boldsymbol{r}_1,\ \frac{\partial \boldsymbol{r}_2}{\partial t} = \boldsymbol{r}_2\times\frac{\boldsymbol{e}_3\times\boldsymbol{a}}{\|\boldsymbol{e}_3\times\boldsymbol{v}\|}\times\boldsymbol{r}_2 \\ &\frac{\partial \boldsymbol{r}_3}{\partial t} = \boldsymbol{r}_3\times\frac{\boldsymbol{a}\times(\boldsymbol{e}_3\times\boldsymbol{v})+\boldsymbol{v}\times(\boldsymbol{e}_3\times\boldsymbol{a})}{\|\boldsymbol{v}\times(\boldsymbol{e}_3\times\boldsymbol{v})\|}\times\boldsymbol{r}_3 \\ &\frac{\partial(\boldsymbol{n}_g^{\top}\boldsymbol{r}_1)}{\partial \bar{\boldsymbol{C}}_i} = \frac{N}{T}\frac{\dot{\boldsymbol{b}}}{\|\boldsymbol{v}\|}\boldsymbol{n}_g^{\top}\left(\boldsymbol{I}-\frac{\boldsymbol{v}\boldsymbol{v}^{\top}}{\boldsymbol{v}^{\top}\boldsymbol{v}}\right)+\frac{N^2}{gT^2}\ddot{\boldsymbol{b}}\boldsymbol{r}_1^{\top} \\ &\frac{\partial(\boldsymbol{n}_g^{\top}\boldsymbol{r}_2)}{\partial \bar{\boldsymbol{C}}_i} = \frac{N}{T}\frac{\dot{\boldsymbol{b}}}{\|\boldsymbol{w}_2\|}\boldsymbol{n}_g^{\top}\left(\boldsymbol{I}-\frac{\boldsymbol{w}_2\boldsymbol{w}_2^{\top}}{\boldsymbol{w}_2^{\top}\boldsymbol{w}_2}\right)[\boldsymbol{e}_3]_{\times}+\frac{N^2}{gT^2}\ddot{\boldsymbol{b}}\boldsymbol{r}_2^{\top} \\ &\frac{\partial(\boldsymbol{n}_g^{\top}\boldsymbol{r}_3)}{\partial \bar{\boldsymbol{C}}_i} = \frac{N}{T}\frac{\dot{\boldsymbol{b}}}{\|\boldsymbol{w}_3\|}\boldsymbol{n}_g^{\top}\left(\boldsymbol{I}-\frac{\boldsymbol{w}_3\boldsymbol{w}_3^{\top}}{\boldsymbol{w}_3^{\top}\boldsymbol{w}_3}\right)\left[2\boldsymbol{e}_3\boldsymbol{v}^{\top}-(\boldsymbol{v}^{\top}\boldsymbol{e}_3)\boldsymbol{I}-\boldsymbol{v}\boldsymbol{e}_3^{\top}\right]+\frac{N^2}{gT^2}\ddot{\boldsymbol{b}}\boldsymbol{r}_3^{\top}\end{aligned} \tag{56}$$

where $[\boldsymbol{e}_3]_{\times} = [0,-1,0;\ 1,0,0;\ 0,0,0] \in \mathbb{R}^{3\times 3}$ is the multiplication cross of $\boldsymbol{e}_3$; $\boldsymbol{w}_2 = \boldsymbol{e}_3\times\boldsymbol{v}$, and $\boldsymbol{w}_3 = \boldsymbol{v}\times(\boldsymbol{e}_3\times\boldsymbol{v})$.

4) Derivatives of Time Mapping

The time mapping in (39) is introduced to ensure the positive definite of $T$. The gradient of (39) is

$$\frac{\partial \mathcal{J}}{\partial \mathcal{T}} = \exp(\mathcal{T}) \cdot \frac{\partial \mathcal{J}}{\partial T} \tag{57}$$

### *C. Trajectory Optimization Process for Fixed-wing UAVs*

Finally, the trajectory optimization is transcribed as

$$\textbf{Problem 4}: \min_{\mathcal{P},\mathcal{T}} \ \mathcal{J}(\mathcal{P},\mathcal{T}) = \mathcal{Q}\left[T(\mathcal{T})\right] + \sum \lambda_\sigma \mathcal{I}_\sigma \left\{ \bar{\mathcal{C}}(\mathcal{P}, T(\mathcal{T})), T(\mathcal{T}) \right\} \tag{58}$$

where $\frac{\partial \mathcal{J}}{\partial \mathcal{P}}$ and $\frac{\partial \mathcal{J}}{\partial \mathcal{T}}$ are analytical.

**Problem 4** is a lightweight unconstrained NLP optimization with analytical gradients. Table I compares (58) with the original problem. Owing to the differential flat trajectory representation, optimization variables $(\mathcal{P},\mathcal{T}) \in \mathbb{R}^{3(N-1)+1}$ are compactly parameterized and do not require directly solving all states and controls as in the original problem, significantly reducing the optimization dimensionality. Through solving $\mathcal{P}$ and $T$, original state spaces $\boldsymbol{x}$, $\boldsymbol{u}$ can be calculated by $\mathcal{F}_x$ and $\mathcal{F}_u$ indirectly. Equality constraints (i.e., the dynamic and terminal constraints) are eliminated to avoid the heavy computational burdens of directly optimizing complex nonlinear dynamics. The inequality constraints are relaxed by integral costs, and thus one can choose appropriate penalty weights $\lambda_\sigma$ and threshold factors $\zeta_\sigma$ to ensure the feasibility. In addition, the analytical gradients are given, which enables the direct utilization of gradients to speed up optimization iteration. Therefore, **Problem 4** has a higher solving efficiency compared to the original problem.

TABLE I
COMPARISONS OF PROBLEMS 1 AND 4

| | Problem 1 | Problem 4 |
|---|---|---|
| Variables | $\boldsymbol{x} \in \mathbb{R}^6$, $\boldsymbol{u} \in \mathbb{R}^3$, $T \in \mathbb{R}_+$ | Compactly parameterized by $\mathcal{P} \in \mathbb{R}^{3\times(N-1)}$, $\mathcal{T} \in \mathbb{R}$ |
| Dimension | $9N+1$ | $3(N-1)+1$ |
| Constraint | Constrained | Unconstrained |
| Gradient | Numerical Calculation (e.g., FDM) | Analytical |

The following presents the detailed procedure of trajectory planning and complexity analysis.

1) Trajectory Planning Procedure

**Algorithm 1** shows the differential flatness-based trajectory optimization procedure for fixed-wing UAVs (DFTO-FW). The main steps are:

**Step 1** (line 1). Use 3D-Dubins connections [38] to generate the initial trajectory guess.

**Step 2** (line 2). Problem normalization and temporal-spatial scaling. We normalize all parameters by $\eta_L$ spatially and $\eta_T$ times temporally to avoid the numerical singularity and improve the adaptability for different scenario scales.

**Step 3** (lines 3-6). Generate the initial trajectory using the differential flat trajectory representation $\text{Traj}_{\text{FW}}$.

**Step 4** (lines 7-21). Iterative loop of the optimization process. Compute the values of objective and analytical gradients. Then, the next iteration is consistently searched using the L-BFGS [39] with the given analytical gradients. When $\|\nabla_{(\mathcal{P}^q,\mathcal{T}^q)}\mathcal{J}\| \le \xi$ ($\xi$ is the convergence tolerance) and $\boldsymbol{x}^q(t)$, $\boldsymbol{u}^q(t)$ satisfy all constraints, the optimization converges. Break the iteration.

**Step 5** (line 22). Problem scale recovery.

**Step 6** (line 23). Calculate the original states and controls and output the optimized trajectory.

**Algorithm 1: DFTO-FW Pseudocode**

**Input:** $\boldsymbol{x}_0$, $\boldsymbol{u}_0$, $\boldsymbol{x}_f$, $\boldsymbol{u}_f$, $\boldsymbol{x}_{\max}$, $\boldsymbol{x}_{\min}$, $\boldsymbol{u}_{\max}$, $\boldsymbol{u}_{\min}$, and $\mathcal{O}_{obs}$.
**Parameters:** $N$, $\kappa$, $\varpi$, $\lambda_\sigma$, $\zeta_\sigma$, $R_{safe}$ and $\xi$.
**Output:** $\boldsymbol{x}(t)$, $\boldsymbol{u}(t)$, and $T$.

**Begin**
1 Use 3D-Dubins as the initial guess. $\eta_L \leftarrow$ Dubins length, $\eta_T \leftarrow \eta_L / V_{\max}$
2 Problem scale normalization. $[\hat{L}] = [L]/\eta_L$ and $[\hat{T}] = [T]/\eta_T$
3 $(\boldsymbol{p}_0, \boldsymbol{v}_0, \boldsymbol{a}_0) \leftarrow \mathcal{F}^{-1}(\boldsymbol{x}_0, \boldsymbol{u}_0)$, $(\boldsymbol{p}_f, \boldsymbol{v}_f, \boldsymbol{a}_f) \leftarrow \mathcal{F}^{-1}(\boldsymbol{x}_f, \boldsymbol{u}_f)$ by (17)
4 $\mathcal{P}^0 = [\boldsymbol{p}_1^0, \ldots, \boldsymbol{p}_i^0, \ldots, \boldsymbol{p}_{N-1}^0]^\top \leftarrow$ 3D-Dubins$(\boldsymbol{x}_0, \boldsymbol{x}_f)/\eta_L$, $\mathcal{T}^0 \leftarrow \ln(T^0)$
5 $\mathcal{B}^{-1}$, $\mathcal{D}^0$, $\bar{\mathcal{C}}^0 \leftarrow$ initialized by (26)-(27)
6 $\boldsymbol{x}^0(t) \leftarrow \mathcal{F}_x(\boldsymbol{p}^0, \dot{\boldsymbol{p}}^0)$, $\boldsymbol{u}^0(t) \leftarrow \mathcal{F}_u(\boldsymbol{p}^0, \dot{\boldsymbol{p}}^0, \ddot{\boldsymbol{p}}^0)$ by (8) -(15)
7 Set $q \leftarrow 0$
8 **while true**
9 $\quad q \leftarrow q+1$, $T^q \leftarrow \exp(\mathcal{T}^q)$
10 $\quad \mathcal{D}^q \leftarrow$ updated by (24), and $\bar{\mathcal{C}}^q \leftarrow$ updated by (27)
11 $\quad \boldsymbol{p}^q(t), \dot{\boldsymbol{p}}^q(t), \ddot{\boldsymbol{p}}^q(t), \dddot{\boldsymbol{p}}^q(t) \leftarrow$ calculated by (19)-(20)
12 $\quad \boldsymbol{x}^q(t) \leftarrow \mathcal{F}_x(\boldsymbol{p}^q, \dot{\boldsymbol{p}}^q)$, $\boldsymbol{u}^q(t) \leftarrow \mathcal{F}_u(\boldsymbol{p}^q, \dot{\boldsymbol{p}}^q, \ddot{\boldsymbol{p}}^q)$ by (8)-(15)
13 $\quad \mathcal{Q}, \mathcal{I}_\sigma, \frac{\partial \mathcal{Q}}{\partial T^q}, \frac{\partial \mathcal{I}_\sigma}{\partial \bar{\mathcal{C}}_i^q}, \frac{\partial \mathcal{I}_\sigma}{\partial T^q} \leftarrow$ computed by (31)-(38) and (48)-(56)
14 $\quad \mathcal{J} \leftarrow \mathcal{Q} + \sum_\sigma \lambda_\sigma \mathcal{I}_\sigma$
15 $\quad \nabla_{(\mathcal{P}^q,\mathcal{T}^q)}\mathcal{J} = \mathrm{col}\{\frac{\partial \mathcal{J}}{\partial \mathcal{P}^q}, \frac{\partial \mathcal{J}}{\partial \mathcal{T}^q}\} \leftarrow$ calculated by (43), (46) and (57)
16 $\quad$ **if** $\|\nabla_{(\mathcal{P}^q,\mathcal{T}^q)}\mathcal{J}\| \le \xi$ and $\boldsymbol{x}^q(t)$, $\boldsymbol{u}^q(t)$ satisfy (2), (3), **then**
17 $\quad\quad$ **break while**
18 $\quad$ **else**
19 $\quad\quad (\mathcal{P}^{q+1}, \mathcal{T}^{q+1}) \leftarrow (\mathcal{P}^q, \mathcal{T}^q) - \alpha^* \boldsymbol{H}_q \nabla_{(\mathcal{P}^q,\mathcal{T}^q)}\mathcal{J}$ by L-BFGS
20 $\quad$ **end if**
21 **end while**
22 Problem scale recovery. $[L] = [\hat{L}] \cdot \eta_L$ and $[T] = [\hat{T}] \cdot \eta_T$
23 **return** $\boldsymbol{x}^*(t) \leftarrow \boldsymbol{x}^q(t)$, $\boldsymbol{u}^*(t) \leftarrow \boldsymbol{u}^q(t)$, $T^* \leftarrow T^q$
**End**

2) Algorithm Time Complexity Analysis

TABLE II
TIME COMPLEXITY COMPARISONS

| | | Solving Problem 1 | DFTO-FW |
|---|---|---|---|
| Objective Evaluation | | $O(N)$ | $O(N)$ |
| Gradients | FDM | $O[N \cdot (9N+1)]$ | $O[N \cdot (3N-2)]$ |
| | Analytical | N/A | $O(N)$ |
| Hessian | FDM | $O[N \cdot (9N+1)^2]$ | $O[N \cdot (3N-2)^2]$ |
| | L-BFGS with FDM Gradients | N/A | $O[mN \cdot (3N-2)]$ |
| | L-BFGS with Analytical Gradients | N/A | $O(mN)$ |

Here, we analyze the time complexity of the proposed DFTO-FW algorithm. Table II compares the time complexity between directly solving the original problem and the DFTO-FW. Take the computation time for each trajectory segment as standard. The calculation of integral costs sums up the cost values for each trajectory segment. Therefore, the time complexity of one objective evaluation is proportional to the trajectory segment number $N$, i.e., $O(N)$. Since the trajectory is compactly parameterized by the intermediate waypoints $\mathcal{P}$ and the flight duration $T$, the total dimension of optimization variables for the DFTO-FW is $3(N-1)+1$. Owing to the derivation of analytical gradients, one can obtain the value of gradients in one function evaluation, which only takes linear time complexity $O(N)$, while using finite difference requires square complexity $O[N\cdot(3N-2)]$. Therefore, analytical gradients significantly reduce the computational burdens. During trajectory optimization, the L-BFGS method utilizes pre-calculated $m$ gradients to approximate the Hessian, which brings linear time complexity $O(mN)$ in the iteration (refer to [39], $m=5\sim40$ depending on the dimension of a specific problem). In summary, the proposed DFTO-FW has linear time complexity $O(mN)+O(N)+O(N)$ in each iteration, thus enabling highly efficient trajectory generation for fixed-wing UAVs.

## V. Numerical Simulations

In this section, the developed DFTO-FW is tested by numerical simulations on typical scenarios to validate the feasibility and efficiency. The simulations are implemented with the computation hardware of Ryzen R5-5600G 3.9GHz CPU and 16 GB RAM. The parameters of flight performance and obstacle avoidance constraints are shown in Table III. According to [31], let $\kappa=5$, $\varpi=3$, $\zeta_\sigma=0.05$, and $\xi=10^{-3}$. Sec.V.A analyses trajectory optimization results in a typical penetration scenario. Sec.V.B presents algorithm comparative studies with several typical algorithms (i.e., GPOPS-II [15], and TRF-SCP [20]) to demonstrate the advantages of DFTO-FW. Sec.V.C and Sec.V.D further discuss the settings of trajectory segment numbers $N$ and the integral cost weights $\lambda_\sigma$. The details are as follows.

TABLE III
PARAMETERS SETTING OF CONSTRAINTS

| Parameter | Range | Parameter | Range / Value |
|---|---|---|---|
| $V$ | [30, 40] m/s | $\gamma$ | [-10, 10] deg |
| $n_x$ | [-0.2, 0.2] | $n_y$ | [-0.2, 0.2] |
| $n_z$ | [0.8, 1.2] | $R_{safe}$ | 100 m |

### A. Trajectory Optimization for a Penetration Scenario

Consider a minimum-time trajectory optimization problem for a fixed-wing UAV in a typical penetration scenario with 15 obstacles. The terminal conditions are set as $\boldsymbol{x}_0$=[500 m, 500 m, -200 m, 30.5 m/s, -90 deg, 0 deg]$^\top$, $\boldsymbol{x}_f$=[9500 m, 500 m, -1800 m, 30.5 m/s, 0 deg, 0 deg]$^\top$ and $\boldsymbol{u}_0=\boldsymbol{u}_f$=[0, 0, 1]$^\top$. The settings of algorithm parameters can be referred to in the discussions on Sec.V.C and Sec.V.D.

The trajectory optimization results are shown in Fig.3. The generated trajectory in Fig.3-(a) achieves smooth and collision-free connections between initial and final positions. The states and controls satisfy the corresponding boundary constraints in (2). In Fig.3-(b), it can be found that the flight speed $V$ shows a trend of increasing first, maintaining maximum speed, and then decreasing to reach the target position quickly. Fig.3-(c) shows that the $n_x$ rises during corresponding acceleration or deceleration, and $n_y$ rises with $\chi$ varying to maneuver fast and avoid collisions. In Fig.3-(d), the UAV maintains a safe distance $d_{obs}\geq100$ m from obstacles to avoid collisions.

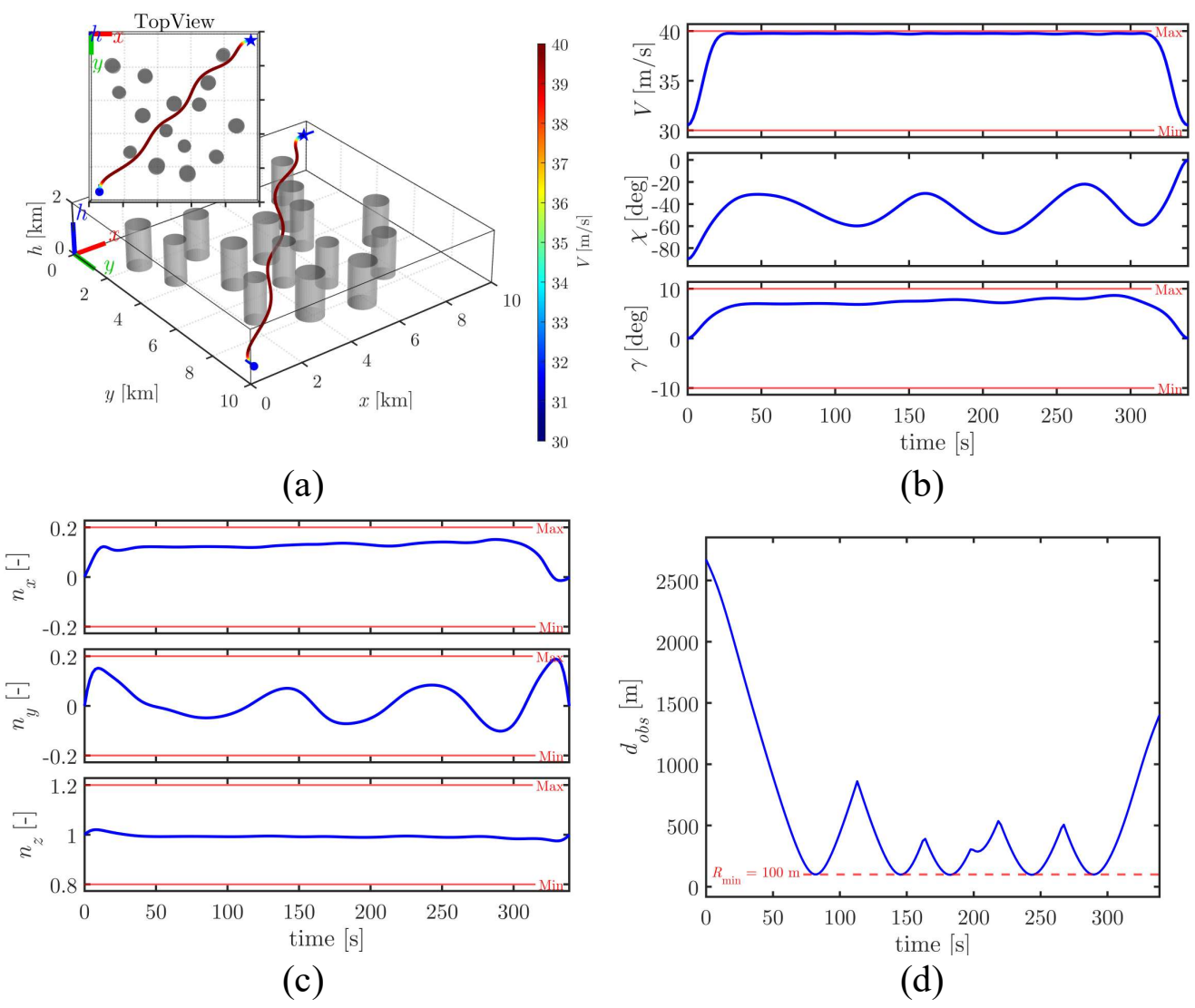

**Fig.3** Optimization results in the penetration scenario. (a) Flight trajectories. (b) States. (c) Controls. (d) Objective and CPU time.

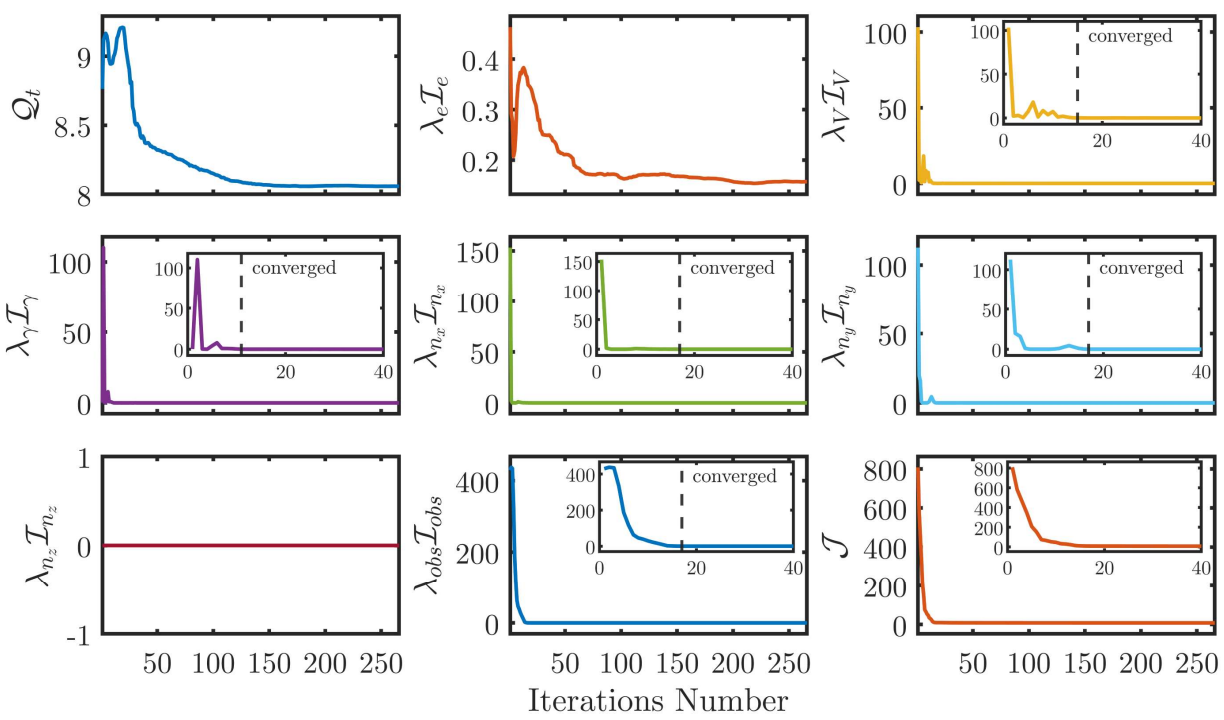

**Fig.4** Convergence history of cost functions.

Fig.4 shows the changes in each cost function with the optimization iterations. As we can see, the values of penalty costs $\lambda_V\mathcal{I}_V$, $\lambda_\gamma\mathcal{I}_\gamma$, $\lambda_{n_x}\mathcal{I}_{n_x}$, $\lambda_{n_y}\mathcal{I}_{n_y}$ and $\lambda_{obs}\mathcal{I}_{obs}$ have large values at the initial iteration, and then rapidly decrease, and they all converge near zero after 20 iterations, which ensures the satisfaction of the flight performance and obstacle avoidance constraints. $\lambda_{n_z}\mathcal{I}_{n_z}$ remains near 0 because the vertical load $n_z$ consistently stays within the feasible range, preventing the penalty from being triggered. It means that DFTO-FW has found a feasible trajectory, which only takes 5.21ms CPU time. Note that, some costs (e.g. $\mathcal{Q}(T)$, $\lambda_e\mathcal{I}_e$, $\lambda_V\mathcal{I}_V$ and $\lambda_\gamma\mathcal{I}_\gamma$) fluctuate and oscillate during the optimization process. It is because certain costs are conflicting. For instance, the flight duration $\mathcal{Q}(T)$ is minimized with increasing speed $V$, which may potentially violate the maximum speed constraint and lead to growing $\mathcal{I}_V$. The optimizer tries to balance these trade-offs and decrease the weighted sum objective in each iteration. After 266 iterations, $\mathcal{J}$ tends towards convergence, and the optimization process satisfies the exit condition, which takes 69.4 ms. The simulation results demonstrate the effectiveness of the proposed methods in engineering practice for the typical penetration scenario.

### *B. Algorithms Comparisons*

To further demonstrate the advantages of the DFTO-FW, this subsection conducts comparative studies with several typical trajectory optimization algorithms for fixed-wing UAVs, i.e., the Radau pseudospectral method solved by GPOPS-II [15] and the TRF-SCP [20] solved by ECOS [40]. The DFTO-FW and the competitors are tested in eight groups of random environments, whose scenario scale for each group is set as (5 + 2.5$i$) km × 5 km× 2 km ($i$ = 1~8) with an obstacle density of 0.4 obstacles/km$^2$. The distance between each group's initial and final positions gradually grows, and the number of obstacles also successively increases (15~50). The locations of obstacles $\boldsymbol{p}_{obs,j}$ are generated by the Latin hypercube sampling method, and the radius $R_{obs,j}$ is random in [200, 400] m. The terminal constraints for each group are set as $\boldsymbol{x}_0$ = [500 m, 2500 m, -500 m, 30 m/s, 0 deg, 0 deg]$^\top$, $\boldsymbol{x}_f$ = [4500+2500$i$ m, 2500 m, -1000 m, 30 m/s, 0 deg, 0 deg]$^\top$ and $\boldsymbol{u}_0$ = $\boldsymbol{u}_f$ = [0, 0, 1]$^\top$. Each group conducts 100 simulations. The settings of other parameters are the same as Sec.V.A.

Fig.5 shows some examples of trajectory optimization results for different algorithms, and Fig.6 summarizes the comparisons of optimality, CPU time, and success rates. It can be seen from Fig.5 that the DFTO-FW generated similar collision-free trajectories to the GPOPS-II and the TRF-SCP. Fig.6-(a)~(b) indicate that the DFTO-FW has comparable optimality with the competitors. At the same time, the CPU time of DFTO-FW is up to about 10$^{-1}$ s, far less than GPOPS-II by two orders of magnitude, and faster than TRF-SCP by one order of magnitude. Especially for the scenario with 50 obstacles, the DFTO-FW only takes an average of 0.14 s to optimize trajectories, far faster than 71.83 s for GPOPS-II and 4.26 s for TRF-SCP. Note that in Fig.6-(a), for the scenarios with 45 and 50 obstacles, the objective values of GPOPS-II and TRF-SCP are abnormally far greater than the DFTO-FW, which is caused by the failure of GPOPS-II and TRF-SCP. In these cases with dense obstacles, GPOPS-II can hardly generate optimal trajectories within limited iterations and CPU time for scenarios with many obstacles. As for TRF-SCP, it is prone to the infeasibility of dynamics convexification, resulting in decreased robustness and computational efficiency. The simulation results demonstrate the advantages of the developed method in terms of high efficiency.

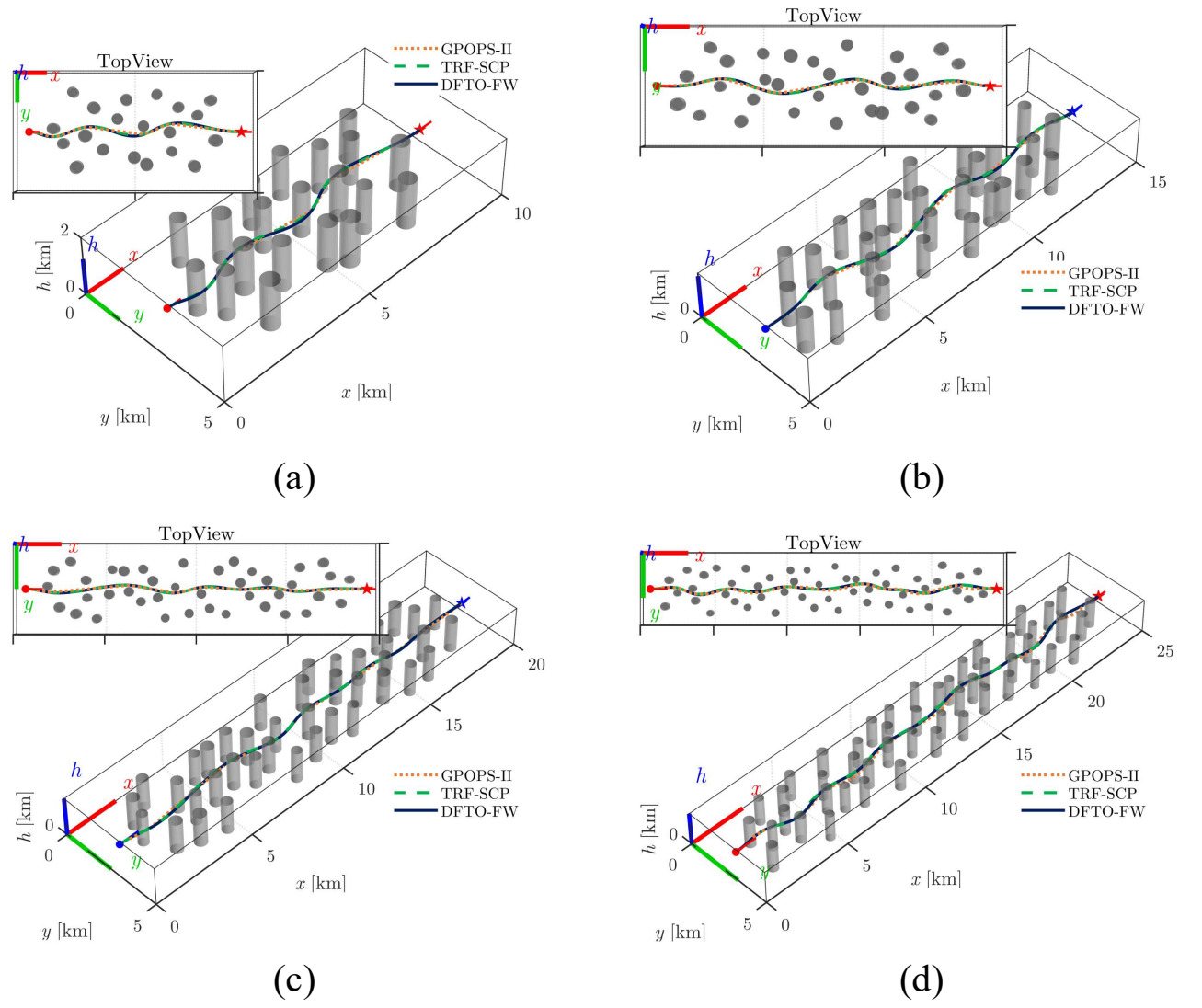

**Fig.5** Trajectory optimization results of comparison algorithms with different obstacle scales. (a) 20. (b) 30. (c) 40. (d) 50.

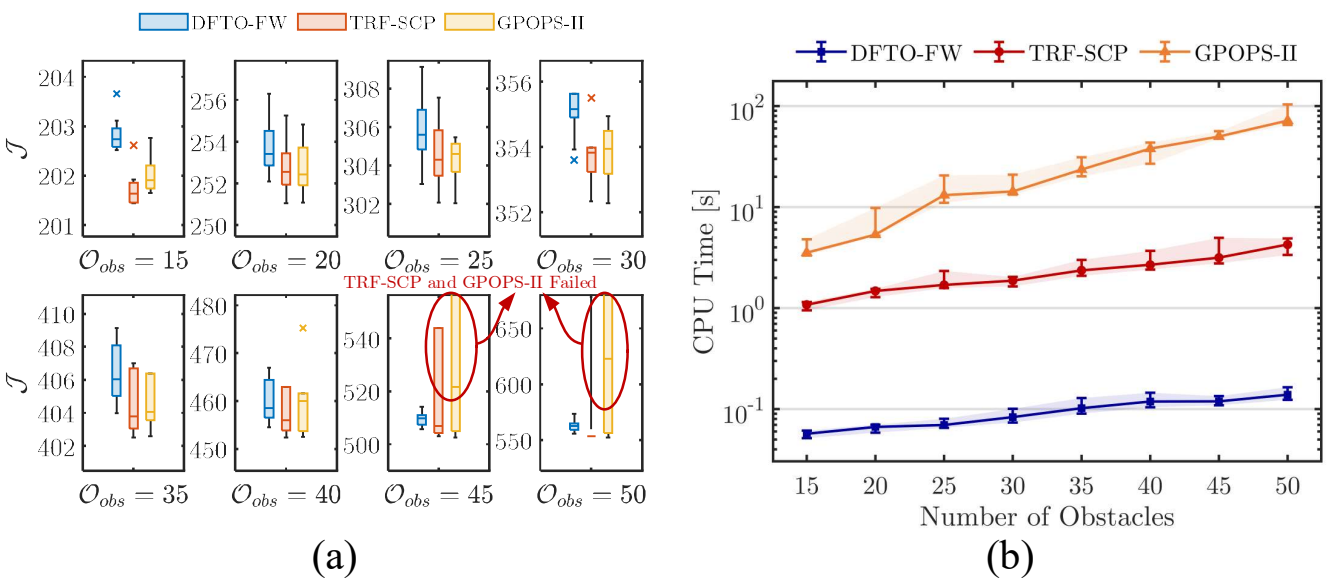

**Fig.6** Algorithms comparison of Monte-Carlo simulations. (a) Optimality. (b) CPU Time.

### *C. Discussion of trajectory segment numbers N*

This subsection discusses the influence of $N$ on the DFTO-FW. We consider a minimum time trajectory optimization with terminal constraints as $\boldsymbol{x}_0$ = [300 m, 4700 m, -500 m, 30 m/s, -90 deg, 0 deg]$^\top$, $\boldsymbol{x}_f$ = [4700 m, 300 m, -1000 m, 30 m/s, -90 deg, 0 deg]$^\top$ and $\boldsymbol{u}_0$ = $\boldsymbol{u}_f$ = [0, 0, 1]$^\top$. There are two cylinder obstacles, where $\boldsymbol{p}_{obs,1}$ = [1800 m, 3800 m]$^\top$, $\boldsymbol{p}_{obs,2}$ = [3200 m, 1200 m]$^\top$ and $R_{obs,1} = R_{obs,2} = 800$ m. Let $\lambda_e$ = 10$^{-3}$, $\lambda_\sigma$ = 10$^3$ and $\zeta_\sigma$ = 0.01 and $\xi$ = 10$^{-3}$. We use the DFTO-FW algorithm to solve the trajectory with different $N$.

The results are shown in Fig.7. As we can see, smaller N (green lines in Fig.7-(a)~(c)) leads to slow acceleration and $V$ cannot increase fast and thus takes longer flight times. It can be also noted that the loads $n_x$ and $n_y$ with smaller $N$ stay far from their boundaries, resulting in slower maneuvers. This phenomenon indicates that a small $N$ will weaken the optimality of the trajectories. Fig.7-(d) depicts the objective values and CPU time concerning N for Case 1. With $N$ increasing (blue deepens in Fig.7-(a)~(c)), $\mathcal{J}$ rapidly decreases, then gradually tends to 167.16 s. This is because $N$ determines the temporal-spatial deformability of $Traj_{FW}$ and further affects the optimality of generated trajectories. However, the CPU time does not significantly increase and rises from 9 ms ( $N$ =5) to 246 ms ( $N$ =25), owing to the linear time complexity of the DFTO-FW with respect to $N$ .

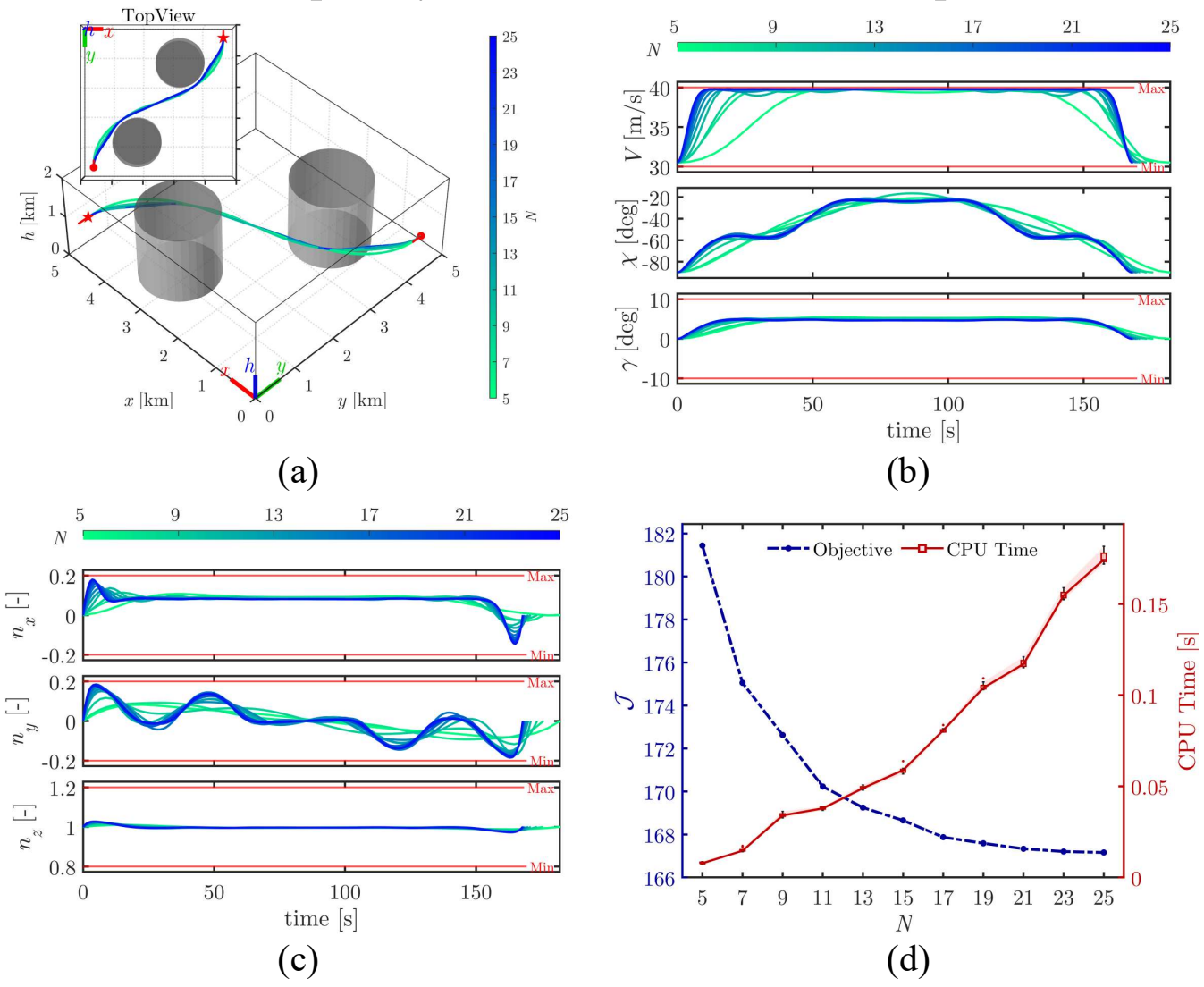

**Fig.7** Trajectory optimization results for different $N$. (a) Flight trajectories. (b) states. (c) Controls. (d) Objective and CPU time.

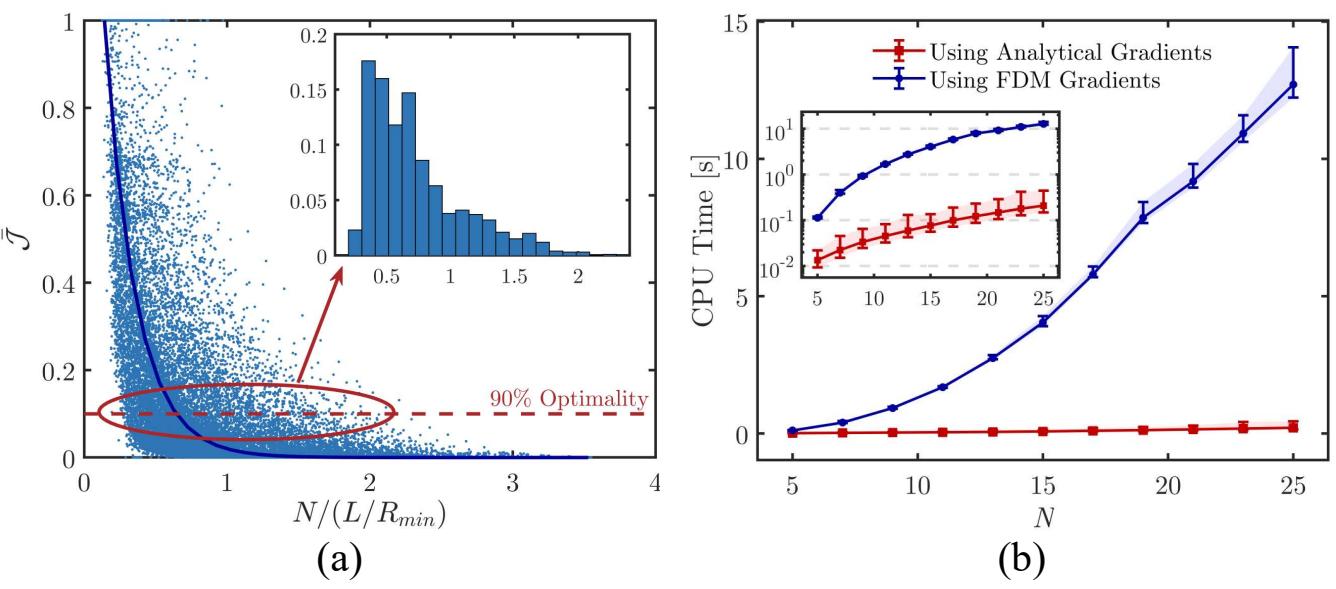

**Fig.8** Objective values and CPU time with respect to $N$. (a) Objective values. (b) CPU time.

To further illustrate the influence of $N$ on optimality and efficiency in general cases, we optimize 1000 different scenarios and record their objective and CPU time. Each scenario has randomly selected initial and final positions and ten random cylinder obstacles in a 10 km × 10 km× 2 km. We solve the trajectory using the DFTO-FW algorithm with different $N$. Fig.8-(a) shows the tendency of the objective with $N$. Since the DFTO-FW uses piecewise polynomials to parameterize the trajectory of fixed-wing vehicles, $N$ should be related to the number of turning maneuverability. Thus, the $x$-axis is normalized by $L/R_{\min}$ ( $L$ is the trajectory length and $R_{\min}=\frac{V_{\min}^2}{g n_{y,\max}}$ is the minimum turning radius). The $y$-axis of Fig.8-(a) is normalized by $\bar{\mathcal{J}}=\frac{\mathcal{J}-\mathcal{J}_{\min}}{(\mathcal{J}_{\max}-\mathcal{J}_{\min})}$. The data of $\bar{\mathcal{J}}$ is fitted as the blue curve, which appears to be a trend of rapid decline first and then slow convergence. Fig.8-(a) hints at how to choose $N$ for DFTO-FW. The sub-figure in Fig.8-(a) captures the distribution of $\frac{N}{L/R_{\min}}$ with $\bar{\mathcal{J}}=0.1$. Refer to the fitted curve, roughly setting $N=\text{round}\left[k_n\cdot\frac{L}{R_{\min}}\right]$, $k_n=1\sim1.5$ is a good choice since nearly 81.1%~95.7% of the samples reached at least 90% of the optimal value. In practice, $L$ can be estimated by the length of 3D-Dubins connections. Fig.8-(b) shows the computation times concerning $N$ in random obstacle environments. Independent of using FDM gradients, we provide analytical expressions of gradients for DFTO-FW to speed up trajectory optimization, dramatically decreasing computation time by an order of magnitude from $10^{-1}$–$10^{1}$ s to $10^{-2}$–$10^{-1}$ s.

### *D. Discussion of Penalty Weights*

Since the core of DFTO-FW is to solve a weighted unconstrained optimization problem, this subsection discusses how to select appropriate weights. The integral costs in (58) can be classified into two major categories: the control effort and penalty costs. We first analyze the influence of penalty weights. Note that the penalties designed in (35)-(38) have been normalized by the corresponding boundary values; thus, they have the same numerical magnitude. Therefore, taking the obstacle avoidance penalty as an example, select different weights $\lambda_{obs}$ to optimize the trajectory for DFTO-FW.

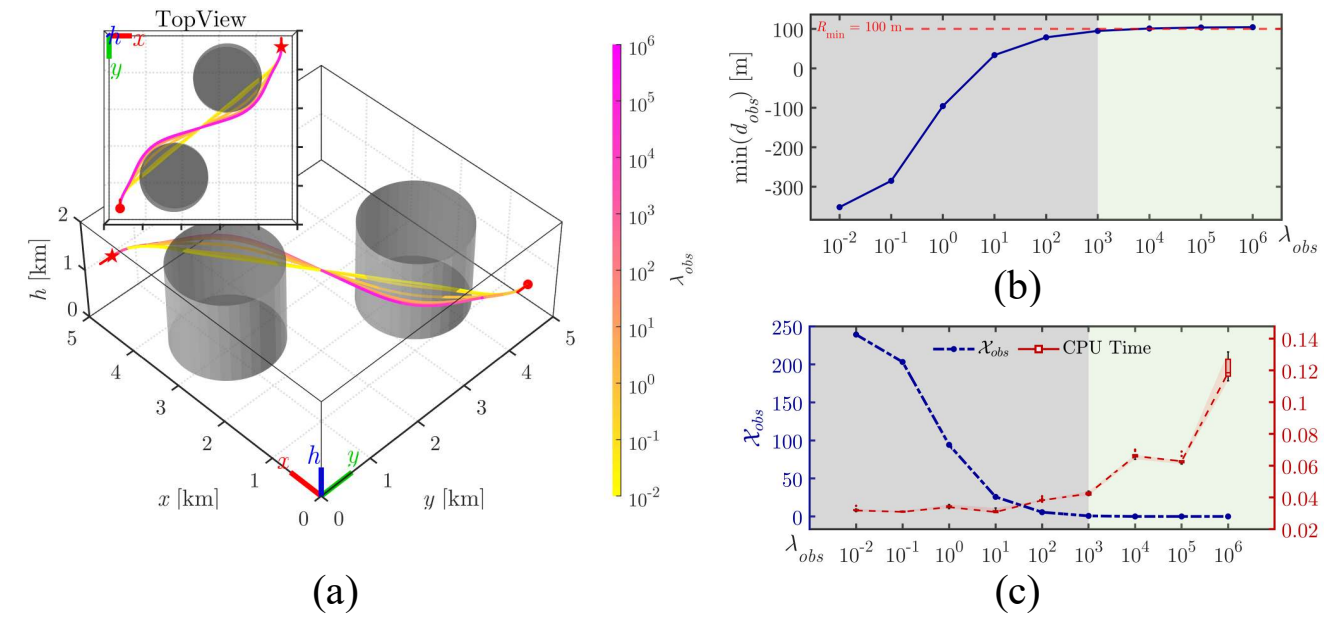

**Fig.9** Optimization results for different $\lambda_{obs}$. (a) Flight trajectory. (b) Distance to obstacles. (c) Constraint violation and CPU time.

Fig.9 shows the simulation results. It can be found that choosing small $\lambda_{obs}$ results in failure of obstacle avoidance, while selecting a large enough penalty weight can guarantee the feasibility of generated trajectories. Fig.9-(c) describes the constraint violation (blue line) and CPU time (red line) for different $\lambda_{obs}$ , where the constraint violation is defined as $\mathcal{X}_{obs}=\int_{t0}^{tf}\max\left[1-d_{obs}/R_{safe},0\right]dt$ . $\mathcal{X}_{obs}$ gradually decreases with increasing $\lambda_{obs}$ . When $\lambda_{obs}\geq10^3$ , $\mathcal{X}_{obs}$ is close to zero, and thus the obstacle avoidance constraint is satisfied. Fig.9-(c) also indicates that too large $\lambda_{obs}$ will take more time to minimize the penalty rather than optimize the original objective. Therefore, we can set $\lambda_\sigma=10^3\sim10^4$ with $\sigma\in\{obs, V,\gamma,n_x,n_y,n_z\}$ to view the balance of feasibility and efficiency.

As for the control effort weight $\lambda_e$, since $\mathcal{I}_e$ is introduced to minimize jerk and smooth the trajectory, $\lambda_e$ is essentially an adjustable parameter to balance time-related cost and trajectory smoothness. Fig.10 shows the simulation results for different $\lambda_e$. Fig.10-(a) visually indicates that the trajectories with larger $\lambda_e$ (yellow lines) are smooth, while the ones with small $\lambda_e$ (green lines) are generally straighter. Fig.10-(b) illustrates the regulating ability of $\lambda_e$ for $\mathcal{Q}=T$ and $E=\int_{t0}^{tf}\boldsymbol{j}^\top\boldsymbol{j}dt$ . When $\lambda_e$ increases, $\mathcal{Q}$ enlarges, and $E$ decreases. Fig.10-(c) shows the oscillation phenomenon of generated trajectories for DFTO-FW. Setting $\lambda_e=0$ (the dashed line) is prone to local oscillations of the polynomial trajectory. Therefore, it is necessary to introduce a small $\lambda_e$, e.g., $\lambda_e = 10^{-3}$ (the solid line), to smooth the trajectory and ease the oscillation phenomenon.

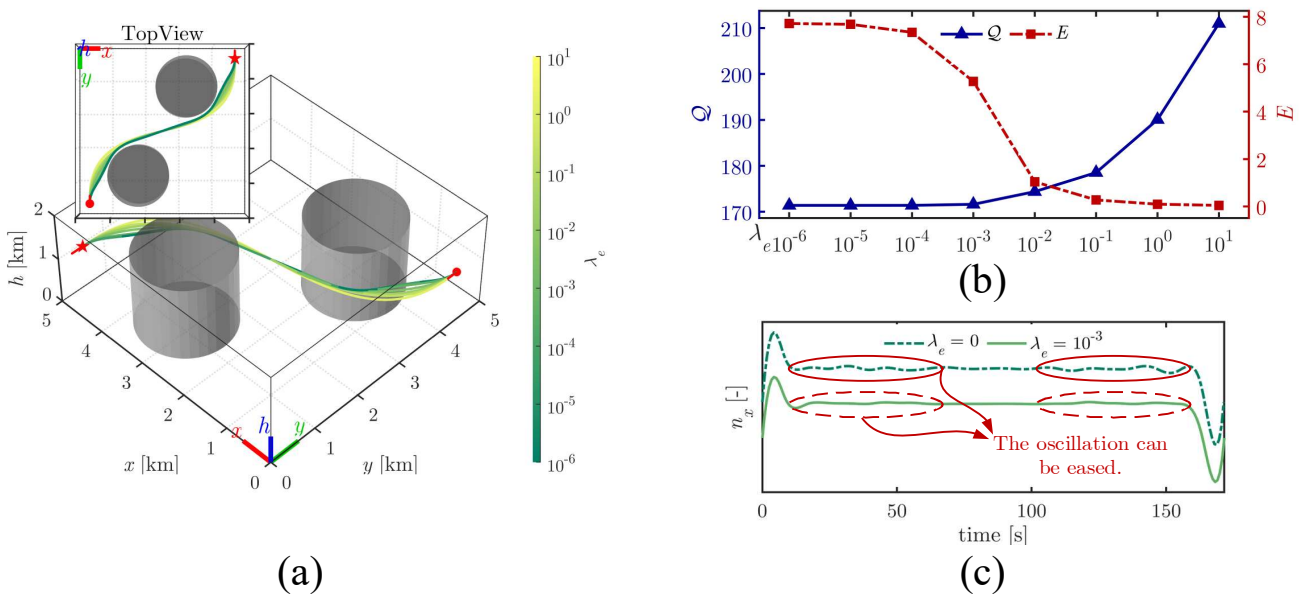

**Fig.10** Trajectory optimization results for different $\lambda_e$. (a) Flight trajectory. (b) Time-related cost and energy effort. (c) Oscillation of trajectories.

## VI. Conclusions

This paper investigates the differential flatness-based trajectory optimization method for fixed-wing UAVs (DFTO-FW). The corresponding optimal control problem is a constrained optimal control problem with complex nonlinear dynamics, making it difficult to solve efficiently using traditional methods. Based on differential flat theory, our approach transcribes the original problem into a lightweight, unconstrained, gradient-analytical optimization with linear time complexity. First, we analyze the differential flatness properties of 3D variable-speed motion of fixed-wing vehicles and then present the customed trajectory representation, which eliminates the computation burdens of satisfying complex nonlinear equality dynamic constraints. Then, the optimization problem is transcribed into an unconstrained optimization with the analytical gradients by designing integral costs. Finally, the DFTO-FW algorithm is proposed to solve the trajectory optimization problem efficiently.

The proposed DFTO-FW has linear time complexity in each optimization iteration. The simulation results illustrate that the proposed method provides superior efficiency, which only takes a sub-second CPU time (on a personal desktop) to generate flight trajectories for fixed-wing UAVs in random obstacle environments. Specifically, in tested scenarios with 50 obstacles, DFTO-FW generates trajectories in an average of 0.14 s with a success rate of nearly 100%, demonstrating a significant advantage in both efficiency and success rate compared to the other competitors.

It should be highlighted that this paper proposes a trajectory representation for fixed-wing UAVs, which helps eliminate the complex dynamics, thus avoiding the heavy computational burdens associated with directly optimizing trajectories based on these dynamics. In addition, our approach derives the analytical gradients for fixed-wing UAV trajectory optimization, accelerating the iteration process and dramatically reducing computation time by an order of magnitude, from $10^{-1}$–$10^{1}$ s to $10^{-2}$–$10^{-1}$ s (when using finite difference calculations). Such a specialized trajectory optimization technique may be extended in other complex nonlinear dynamic systems.